\documentclass[preprint, 12pt]{elsarticle}
\usepackage{url}
\usepackage{graphicx}
\usepackage{algorithmic}
\usepackage{lscape}
\usepackage{float}
\journal{Artificial Intelligence}

\Large
\normalsize

\date{}

\begin{document}

\title{Evaluation of a Simple, Scalable, Parallel Best-First Search Strategy} %
\author[titech]{Akihiro Kishimoto}
\ead{kishimoto@is.titech.ac.jp}
\address[titech]{Tokyo Institute of Technology}
\author[ut]{Alex Fukunaga}
\address[ut]{The University of Tokyo}
\ead{fukunaga@idea.c.u-tokyo.ac.jp}
\author[ibm]{Adi Botea\footnote{Part of this research was performed
while Adi Botea was affiliated with NICTA and The Australian National University.}}
\ead{adibotea@ie.ibm.com}
\address[ibm]{IBM Research, Dublin, Ireland}


\begin{abstract}
Large-scale, parallel clusters composed of commodity processors are
increasingly available, enabling the use of vast processing
capabilities and distributed RAM to solve hard search problems.
We investigate Hash-Distributed A* (HDA*), a simple approach to parallel best-first search
that asynchronously distributes and schedules work among
processors based on a hash function of the search state.
We use this approach to parallelize the A* algorithm in an
optimal sequential version of the Fast Downward planner, as well as a 24-puzzle solver.
The scaling behavior of HDA* is evaluated experimentally on
a shared memory, multicore machine with 8 cores, a cluster of commodity machines using up to 64 cores, 
and large-scale high-performance clusters, using up to 2400 processors.
We show that this approach scales well, allowing the effective
utilization of large amounts of distributed memory to optimally solve problems which
require terabytes of RAM.
We also compare HDA* to Transposition-table Driven Scheduling (TDS), a hash-based parallelization of IDA*, and show that, in planning, HDA* significantly outperforms TDS.
A simple hybrid which combines HDA* and TDS to exploit strengths of both algorithms is proposed and evaluated.
\end{abstract}

\maketitle

\section{Introduction}
\label{sec:introduction}

Parallel search is an important research area for two reasons.
First, many search problems, including planning instances, 
continue to be difficult for sequential algorithms.
Parallel search on state-of-the-art, parallel clusters has the potential
to provide both the memory and the CPU resources required to solve
challenging problem instances. 
Second, while multiprocessors
were previously expensive and rare, multicore machines are now
ubiquitous. 
Future generations of hardware are likely to continue to have an
increasing number of processors, where the speed of each individual
CPU core  does not increase as rapidly as in past decades. Thus,
exploiting parallelism is necessary to extract significant
speedups from the hardware.

Our work is primarily motivated by domain-independent planning.
In classical planning, many problem instances continue to pose a challenge for state-of-the-art planning systems.
Both the memory and the CPU requirements are main causes of performance bottlenecks.
The problem is especially pressing in sequential optimal planning.
Despite significant progress in recent years in developing domain-independent admissible heuristics
\cite{Haslum:00,Edelkamp:01,Helmert:07},
scaling up optimal planning remains a challenge.
Multi-processor, \emph{parallel planning}\footnote{In this paper,
parallel planning refers to computing sequential plans with multi-processor planning,
as opposed to computing parallel plans with a serial algorithm.}
has the potential
to provide both the memory and the CPU resources required to solve
challenging problem instances. 

We introduce and evaluate Hash Distributed A* (HDA*),
a parallelization of A* \cite{Hart:68}.
HDA* runs A* on every processor, where each processor has its own open and closed lists.
A hash function assigns each state to a unique processor, so that every state has an ``owner''.
Whenever a state is generated, its owner processor is computed according to this hash function, and the state is sent to its owner.
This simple mechanism simultaneously accomplishes load balancing as well as duplicate pruning.
While the  key idea of hash-based assignment of states to processors
was initially proposed as part of the PRA* algorithm by Evett et al. \cite{Evett:1995}, %
and later extended by Mahapatra and Dutt \cite{MahapatraD97}, the scalability 
and limitations of hash-based work distribution and duplicate pruning
have not been previously evaluated in depth. 
In addition, hash-based work distribution has never been applied to domain-independent planning.

HDA* has two key attributes which make it worth examining in detail.
First, it is inherently scalable on large-scale parallel systems, because it is a distributed algorithm with no central bottlenecks.
While there has been some recent work in parallel search
\cite{KorfS05,Zhou:07,Burns:09}, these approaches are multi-threaded, and
limited to a single, shared-memory machine. %
HDA*, on the other hand, scales naturally from a single, multicore desktop machine to a large scale, distributed memory cluster of multicore machines.
When implemented using the standard MPI message passing library \cite{SnirG98}, the
exact same code can be executed on a wide array of parallel
environments, ranging from a standard desktop multicore to a massive
cluster with thousands of cores, effectively using all of the aggregate CPU and memory resources available on the system.

Second, HDA* is a very simple algorithm -- conceptually, the only
difference between HDA* and A* is that when a node is generated, we
compute its hash value, and send the node to the closed list of the processor that ``owns'' the
hash value.  Everything runs asynchronously, and there is no tricky synchronization.
This simplicity is extremely valuable in parallel algorithms, as parallel programming is notoriously difficult.
Thus, the goal of our work is to evaluate the performance and the scalability of HDA*.

HDA* is implemented as an extension of  two state-of-the-art solvers.
The first solver is the Fast Downward domain-independent planner.
We use the cost-optimal version with the explicit (merge-and-shrink) abstraction heuristic
reported by  Helmert, Haslum, and Hoffman \cite{Helmert:07}.
The second solver is an application-specific 24-puzzle solver, which uses the pattern database heuristic code provided by Korf and Felner \cite{KorfF02}.
A key difference between these is the relative speed of processing an individual state. 
The domain-specific 24-puzzle solver processes states significantly faster than 
the domain-independent Fast Downward planner, expanding 2-5 times more states per second.\footnote{Korf and Felner's code uses IDA*, which generates states much faster; we only incorporate their pattern database heuristic code in our parallel A* framework.}
The speed of processing a state can significantly impact the \emph{efficiency} of 
a parallel search algorithm, which is the speedup relative to a serial implementation 
divided by the number of CPU cores.
A larger processing cost per state tends to 
diminish the impact of parallel-specific overheads,
such as the communication and the synchronization overhead
(introduced in Section \ref{sec:background}).
Therefore, using both types of solvers allows us to assess
the efficiency of HDA* across a range of
application problems.

The scaling behavior of the algorithm is evaluated
on a wide range of parallel machines.
First, we show that on a standard, 8-core workstation, HDA* achieves speedups up to 6.6.
We then evaluate the exact same HDA* code on a commodity cluster with an Ethernet network, and two high-performance computing (HPC) clusters with up to 2400 processors and 10.5TB of RAM.
The scale of our experiments goes
well beyond previous studies of hash-based work distribution. %
We show that HDA* scales well, allowing
effective utilization of a large amount of distributed
memory to optimally solve problems that require 
terabytes of RAM.

The experiments include a comparison with TDS~\cite{Romein:99,Romein:2002},
a successful parallelization of IDA* with a distributed transposition table.
Using up to 64 processors on a commodity cluster, we show that HDA* is 
2-65 times faster and solves more instances than TDS.
We propose a simple, hybrid approach that combines 
strengths of both HDA* and TDS:
Run HDA* first, and 
if HDA* succeeds, return the solution. 
Otherwise, start a TDS search where the initial threshold for the depth-first exploration
is provided by the failed HDA* search.
Clearly, when HDA* succeeds, this hybrid algorithm runs as fast as HDA*.
We show that, when HDA* fails, the total runtime of the hybrid approach
is comparable to the running time of TDS.

The rest of the paper is organized as follows.
Section \ref{sec:background} presents background information.
The HDA* algorithm is described in Section \ref{sec:hda*}.
We then present an empirical evaluation and analysis of the scalability HDA* in Section \ref{sec:results-scaling}.
Tuning the performance of HDA* is addressed in Section~\ref{tuning-hda}.
Hash-based distribution is compared with a simpler, randomized work distribution strategy in Section \ref{sec:hda-random-distribution-results}.
In Section \ref{sec:hda*-vs-tds}, we compare HDA* with TDS, and propose a hybrid strategy which combines the strengths of both algorithms.
We review other approaches to parallel search in Section \ref{sec:related-work}.
This is followed by a summary and discussion of the results, and directions for future work.

\section{Background }
\label{sec:background}

Efficient implementation of parallel 
search algorithms is challenging %
due to several types of overhead.
{\it Search overhead} occurs when 
a parallel implementation of a search algorithm expands (or generates)
more states than a serial implementation.
The main cause of search overhead is partitioning of the search space
among processors,
which has the side effect that the access to non-local information is restricted.
For example, sequential A* can terminate immediately after
a solution is found, because it is guaranteed to be optimal. 
In contrast, when a parallel A* algorithm finds a (first)
solution at some processor, it is not necessarily a globally  optimal solution.
Better solutions might exist in non-local portions of the search space.
A more detailed discussion of the search overhead is available in Section~\ref{so-planning}.

{\it Synchronization overhead} is the idle time wasted at synchronization points,
where some processors have to wait for the others to
reach the synchronization point.
For example, in a shared-memory environment, the idle time can be
caused by mutual exclusion locks on shared data.
Finally, {\it communication overhead} refers to the cost of 
inter-process information exchange in a distributed-memory environment (i.e., the cost of sending a message from one processor to another over a network).

The key to achieving a good speedup in parallel search is to minimize such overheads.
This is often a difficult task, in part because the overheads are interdependent.
For example, reducing search overhead usually increases synchronization and communication
overhead.

There are several, broad approaches to parallelizing search
algorithms.
This paper focuses on parallelization by partitioning the search
space, so this section reviews this approach.
Other approaches, including 
parallelization of the computation performed on each state, 
running a different search algorithm on each processor, and
parallel window search,
 are reviewed in Section \ref{sec:related-work}.
In addition, this paper focuses on parallel best-first search. The use of hash-based work distribution techniques similar to HDA* for breadth-first search in model checking is reviewed in Section~\ref{sec:related-work}.

One general framework for  partitioning the search space among
parallel processes 
first starts with a {\em root process} which initially generates some seed nodes 
using sequential search, and assigns these seed nodes among the available
processors. Then, at each processor, a search algorithm begins to explore the descendants of its seed node.
These seed nodes become the local root nodes for a local, sequential search algorithm.
This basic strategy can be applied to depth-first search algorithms,
including simple depth-first search, branch-and-bound, and IDA*, as
well as breadth-first and best-first search algorithms such as A*.

{\it Work stealing} is a standard approach for partitioned, parallel search,
and is used in many applications, particularly in shared-memory environments.
In work-stealing, each processor maintains a local work queue.
When a processor $P$ generates new work (i.e., new states to be expanded) $w$, 
it places $w$ in $P$'s own local queue.
When $P$ has no work in its queue, it ``steals'' work from the queue of a busy processor.

Two key considerations in a particular work-stealing strategy are how
to decide which processor to steal work from, as well as how to decide
which and how much work to steal. Various work-stealing
strategies for depth-first, linear-space algorithms such as
depth-first branch-and-bound IDA*, and minimax search have been
studied (e.g. \cite{Rao:1987,Feldmann:1993,Frigo:98}). While most
work on work-stealing has been on MIMD
systems, parallelization of IDA* on
SIMD machines
using an alternating, two-phase mechanism, with
a search phase and a load balancing phase, has also been investigated
\cite{PowleyFK93,MahantiD93}.\footnote{In MIMD (Multiple-Instruction,
Multiple-Data stream) systems, each processor executes code
independently; in SIMD (Single-Instruction, Multiple-Data stream) systems, all processors execute the same instruction (on different data).} 
Another approach to search space partitioning (particularly in shared-memory search) is derived from a line of work 
on addressing memory capacity limitations by using a large amount of slower, external memory (such as disks), to store states in search \cite{Korf04,Korf08,Zhou:04} 
(external memory was also used specifically for planning \cite{EdelkampJ06}).
An issue with using external memory is the overhead of expensive
I/O operations, so techniques for structuring the search to minimize these overheads have been the focus of work in this area.
Korf has implemented a multithreaded, breadth-first search using a shared work queue which uses external memory \cite{KorfS05,Korf08}.
Interestingly, some approaches to reducing the I/O overhead in external memory search
can be adapted to handle the inter-process communication overhead in parallel search.
Zhou and Hansen \cite{Zhou:07} introduce a parallel, breadth-first search algorithm.
Parallel structured duplicate detection seeks to reduce synchronization overhead.
The original state space is partitioned into collections of states called blocks.
The duplicate detection scope of a state contains the blocks that 
correspond to the successors of that state.
States whose duplicate detection scopes are disjoint can be
expanded with no need for synchronization.
Burns et al. \cite{Burns:09} have investigated best-first search algorithms
that include enhancements such as structured duplicate detection and speculative search.
These techniques were shown to be effective in a shared memory machine with up to 8 cores.

We now review the line of work directly related to HDA*.
Algorithms such as breadth-first or best-first search
(including A*) use an open list which stores the set of states that have been generated but not yet expanded.
In an early study, Kumar, Ramesh, and Rao \cite{Kumar:1988}
identified two broad approaches to parallelizing best-first search, based on how the usage and maintenance of the open list was parallelized.
In a {\em centralized} approach, a single open list is
shared among all processes. Each process expands one of the current
best nodes from the globally shared open list, and generates and evaluates its
children. This centralized approach introduces very little or no 
search overhead, %
and no load balancing among processors is necessary.
Furthermore, this method is especially simple to implement in a
shared-memory architecture by using a shared data structure for the
open list. However, %
concurrent access to the shared
open list becomes a bottleneck and inherently limits the scalability
of the centralized approach, except in cases where the cost of
processing each node (e.g., evaluating the node with a heuristic
function) is extremely expensive, in which case
overheads associated with shared open list access become insignificant.

In contrast, in a {\em decentralized} approach to parallel best-first
search, each process has its own open list.
Initially, the root processor generates and distributes
some search nodes among the available processes. Then, each process
starts to locally run best-first search using its local open
list (as well as a closed list, in case of %
algorithms such as A*). 
Decentralizing the open list eliminates the concurrency overheads associated with a 
shared, centralized open list, but load balancing becomes necessary.

Kumar, Ramesh and Rao
\cite{Kumar:1988}, as well as Karp and Zhang \cite{KarpZ88,KarpZ93}
proposed a random work allocation strategy, where newly generated states were sent to
random processors. 
In parallel architectures with non-uniform communication costs, a straightforward variant of
this randomized strategy is to send states to a random neighboring processor (with low communication cost) to avoid the cost of sending to an arbitrary processor (c.f., \cite{DuttM94}).

In addition to load balancing, another issue that a parallel search
algorithm must address is duplicate detection.
In many search applications, including domain-independent planning,
the search space is a graph rather than a tree, and there are multiple paths to 
the same state. 
In sequential search, duplicates can be detected and pruned
by using a closed list (e.g., hash table) or other
duplicate detection techniques (e.g. \cite{Korf:2000,Zhou:2006}).
Efficient duplicate detection is critical for performance,
both in serial and parallel search algorithms, and 
can potentially eliminate vast amounts of redundant work.

In parallel search, duplicate state detection %
incurs several overheads,
depending on the algorithm and the machine environment.
For instance, in a shared-memory environment, 
many approaches, including work-stealing,
need to carefully manage locks on the shared open and closed lists.

Parallel Retracting A* (PRA*) \cite{Evett:1995}
simultaneously addresses the problem of work distribution and duplicate state detection.
In PRA*, each processor maintains its own open and closed lists.
A hash function maps each state to exactly one processor which ``owns'' the state.
When generating a state, PRA* distributes it to the corresponding owner.
If the hash keys are distributed uniformly across the processors, 
load balancing is achieved.
After receiving states, PRA* has the advantage that duplicate detection can be performed efficiently and locally at the destination processor.

While PRA* incorporated the idea of hash-based work distribution, PRA*
differs significantly from a parallel A* in that it is a parallel
version of RA* \cite{Evett:1995}, a limited memory search algorithm closely related to MA*
\cite{ChakrabartiGAS89} and SMA* \cite{Russell92}.  When a processor's
memory became full, Parallel Retracting A* {\em retracts} states from
the search frontier, and their $f$-values are stored in their parents,
which frees up memory.  Thus, unlike parallel A*, PRA* does not store
all expanded nodes in memory, and will not terminate due to running
out of memory in some process.  On the other hand, 
the implementation of this retraction mechanism in \cite{Evett:1995} incurs
a significant synchronization overhead:
when a processor $P$ generates a
new state $s$ and sends it to the destination processor $Q$, $P$
blocks and waits for $Q$ to confirm that $s$ has successfully been
received and stored (or whether the send operation failed due to
memory exhaustion at the target process).
Unlike PRA*, HDA* does not incorporate a node retraction mechanism. 
Also, unlike the original implementation of PRA*, 
HDA* is a fully asynchronous algorithm, where all messages are sent/received asynchronously.

The idea of hash-based work distribution was investigated further by
Mahapatra and Dutt \cite{MahapatraD97}, who studied parallel A* on a
Hypercube architecture, where CPUs are connected by a hypercube network,
while in current standard architectures machines are connected by either a mesh or
torus network.
As a baseline, they implemented Global Hashing (GOHA), which is similar to PRA*, except that
GOHA is a parallelization of SEQ\_A*, a variant of A* which performs partial
expansion of states, while 
HDA* is a parallelization of standard A*.
They proposed two alternatives to the simple
hash-based work distribution in PRA*.  The first approach by Mahapatra
and Dutt, called Global Hashing of Nodes and Quality Equalizing
(GOHA\&QE), decouples load balancing and duplicate checking. States
are assigned an owner (based on hash key) for duplicate checking, and
a newly generated state is first sent to its owner process for
duplicate checking. If the state is in the open list of the owner
process, then it is a duplicate and discarded. Otherwise, it is added
to the open list of the owner, and the state is possibly
reassigned to another process using Dutt and Mahapatra's Quality
Equalizing (QE) strategy \cite{DuttM94}.
In both PRA* and GOHA\&QE, the hash function is global -- a state can
be hashed to any of the processors in the system. In parallel
architectures where communication costs vary among pairs of
processors, such a global hashing may be suboptimal. Therefore,
Mahapatra and Dutt also proposed Local Hashing of Nodes and QE
(LOHA\&QE), which incorporates a state space partitioning strategy and
allocates disjoint partitions to disjoint processor groups in order to
minimize communication costs \cite{MahapatraD97}.
Mahapatra and Dutt showed that GOHA\&QA and LOHA\&QE outperformed the
simpler, global hash-based work distribution method used in PRA* on
the Travelling Salesperson Problem (TSP).

Mahapatra and Dutt's local hashing is based on a number of restrictive 
assumptions that do not allow applying this strategy to planning.
Specifically, local hashing works in problems
with so-called levelized search graphs. 
In a levelized graph, a given state will always have the same depth (distance from root node),
regardless of the path that connects the root and the state. 
The notion of the levelized graph can sometimes be extended to exploit local hashing if
the search space has some regularities on depths such as multiple sequence alignment \cite{KobayashiKW2011}. 
However, planning and the sliding-tile puzzle do not belong to this class of problems.
The reason is that the same state could be reached via different-length paths.
For example, if two cities A and B are connected by two routes of different lengths,
then driving a truck from A to B via each route will result in the same state but the
paths from the starting state will have different lengths.

Transposition-table driven work scheduling (TDS)~\cite{Romein:99,Romein:2002}
is a distributed memory, parallel IDA* algorithm.
Similarly to PRA*, TDS distributes work using a state hash function.
The transposition table is partitioned over processors to be used
for detecting and
pruning duplicate states that arrive at the processor. 
Thus, TDS distributes a transposition table for IDA* among the processing nodes, similarly to how PRA* distributes 
the open and the closed lists for A*.
This distributed transposition table allows TDS to exhibit a very low (sometimes negative) search overhead,
compared to a sequential IDA* that runs on a single computational node 
with limited RAM capacity.
TDS achieved impressive speedups in applications such as the 15-puzzle, 
the double-blank puzzle, 
and the Rubik's cube, on a distributed-memory machine.
The ideas behind TDS have also been successfully integrated in %
adversarial two-player search \cite{Kishimoto:02,Romein:2003, YoshizoeKKYI2011}. 

Thus, the idea of hash-based distribution of work is
not new, but there are several reasons to revisit the idea and perform
an in-depth evaluation at this point. 
First, the primary motivation
for this work was to advance the state of the art of 
domain-independent planning by parallelizing search. While there has been some previous,
smaller-scale work on parallel planning, large-scale parallel planning
has not been previously attempted, and hash-based
work distribution is a natural approach for scaling parallel planning
to large-scale parallel clusters.

Second, parallel systems have
become much more common today than when the earlier work by Evett et
al. \cite{Evett:1995} and Mahapatra and Dutt \cite{MahapatraD97} was
done, and the parallel systems which are prevalent today have very
different architectures. The most common parallel architectures today
are commodity, multicore, shared memory machines, as well as
distributed memory clusters which are composed of shared memory
multicore nodes. 
Hash-based work distribution is a simple approach that can
potentially scale naturally from single, multicore nodes to large
clusters of multicore nodes, and it is important to evaluate its
performance on current, standard parallel architectures.  
In addition, previous algorithms based on this idea, such as PRA* and Mahapatra and Dutt's method,
make some assumptions that are specific to the hardware architecture in use.
In contrast, we aim at obtaining an algorithm as general as possible, avoiding hardware-specific assumptions. 
In fact, as mentioned earlier, our MPI-based implementation can be run on a variety of platforms,
including shared-memory and distributed-memory systems.

Third,
there are some important issues that were not fully explored in the earlier
work. For example, factors which can potentially limit the scaling of hash-based work distribution, such as search overhead and communication overhead, have not been analyzed in detail. 
Also, the performance impact of asynchronous vs synchronous communication, as well as the 
impact of using a hash function
for work distribution, as opposed to a randomized strategy,
have not been previously investigated.  

Fourth, while the work on TDS showed the utility of asynchronous,
hash-based work distribution for IDA*, this previous work was done on
15-puzzle variants and the Rubik's cube, which are two domains where
the overhead incurred by re-exploration of states  in IDA* is known to be
relatively small. In some other domains, this overhead can be quite
significant, which can result in significant costs on a cluster
environment.  Thus, an investigation of the scalability of hash-based,
parallel A* and a comparison with TDS is worthwhile.

Thus, while previous work has considered global hash-based
distribution either as a component of a more complex algorithm
\cite{Evett:1995} or as a straw man against which local hash-based
distribution was considered \cite{MahapatraD97}, this is the first
paper which analyzes the scalability and limitations of global hashing
in depth.
An early version of this work has been previously  presented in a conference paper~\cite{ICAPS09}. However, that initial work
was limited to up to 128 CPU cores on an older system, contained no results on the 24-puzzle, 
no comparison to TDS,
and included a less detailed evaluation and analysis.

 \section{Hash Distributed A*}
\label{sec:hda*}

We now describe Hash Distributed A* (HDA*), a simple
parallelization of A* which uses the hash-based work
distribution strategy originally proposed in PRA* \cite{Evett:1995}. 
In HDA* the closed and open lists are implemented as a distributed
data structure, where each processor ``owns'' a partition of the entire search space.
The local open and closed list for processor $P$ is denoted $Open_P$ and $Closed_P$.
The partitioning is done via a hash function on the state, as described later.

HDA* starts by expanding the initial state
at the root processor.
Then, each processor $P$ executes the following loop until an optimal solution is found: 
\begin{enumerate}
\item
First, $P$ checks if one or more
new states have been received in its message queue. If so, $P$
checks for each new state $s$ in $Closed_P$, in order to
determine whether $s$ is a duplicate, or whether it should be
inserted in $Open_P$\footnote{Even if the heuristic function \cite{Helmert:07} is consistent,
parallel A* search may sometimes have to re-open a state saved in the closed list.
For example, $P$ may receive many identical states with various priorities 
from different processors and these states may reach $P$ in any order.}.

\item
If the message queue is empty, then $P$ selects a highest priority
state from $Open_P$ and expands it, resulting in newly generated states.
For each newly generated state $s$, a hash key $K(s)$ is computed based on the state representation, and $s$
is sent to the processor which owns $K(s)$. This send is asynchronous and non-blocking. $P$ continues its computation without waiting for a reply from the destination.
\end{enumerate}

In a straightforward implementation of hash-based work
distribution on a shared memory machine, each thread
owns a local open/closed list implemented in shared memory,
and when a state $s$ is assigned to some thread, the writer thread obtains
a lock on the target shared memory, writes $s$, then releases
the lock.  Note that whenever a thread $P$ ``sends'' a state $s$ to
a destination $dest(s)$, then $P$ must wait until the lock for
shared open list (or message queue) for $dest(s)$ is available and not
locked by any other thread.  This results in significant
synchronization overhead -- for example, it was observed in
\cite{Burns:09} that a straightforward implementation of PRA*
exhibited extremely poor performance on the Grid search problem, 
and multicore performance for up to 8 cores was consistently {\em slower}
than sequential A*.
While it is possible to speed up locking operations by using, for
example, highly optimized lock operations implementations in inline
assembly language, %
the performance degradation due to synchronization %
remains a considerable problem.

In contrast, the open/closed lists in HDA* are not explicitly shared among the processors. 
Thus, even in a multicore environment where it is possible to share memory, all communications are done 
between separate MPI processes using non-blocking send/receive operations.
Our program implements this by using {\tt MPI\_Bsend} and {\tt MPI\_Iprobe},
and relies on highly optimized message buffers
implemented in  MPI.

Every state must be sent from the processor where it is generated to its ``owner'' processor.
In their work with transposition-table driven scheduling for parallel IDA*, 
Romein et al. \cite{Romein:99} showed that this communication overhead could be overcome by packing multiple 
states with the same destination into a single message.
HDA* uses this state packing strategy to reduce the number of messages.
The relationship between performance and message sizes depends on several factors such as network configurations, the number of CPU cores, and CPU speed.
In our experiments, 100 states are packed into each message on a commodity cluster using
more than 16 CPU cores and a HPC cluster, while 10 states are packed on the commodity cluster
using less than 16 cores. 

In a decentralized parallel A* (including HDA*), when a solution is discovered,
there is no guarantee at that time that the solution is optimal~\cite{Kumar:1988}.
When a processor discovers a locally optimal solution, the processor broadcasts its cost.
The search cannot terminate until all processors have proved that there is no
solution with a better cost.
In order to correctly terminate a decentralized parallel A*, it is not sufficient to check the local open list at every processor.
We must also ensure that there is no message en route to some processor that could lead to a better solution.
Various algorithms to handle termination exist. In our implementation of HDA*, we used the 
time algorithm of Mattern \cite{Mattern:87}, which was also used in TDS.

Mattern's method is based on counting sent messages and received messages.
If all processors were able to count simultaneously, %
it would be trivial to detect whether a message is still en route.
However, in reality, different processors $P_i$ will report their sent and received counters,
$S(t_i)$ and $R(t_i)$, at different times $t_i$.
To handle this, Mattern introduces a basic method where the counters are reported in two different waves.
Let $R^* = \sum_i R(t_i)$ be the accumulated received counter at the end of the first wave,
and $S'^* = \sum_i S(t'_i)$ be the accumulated sent counter at the end of the second wave.
Mattern proved that if $S'^* = R^*$, then the termination condition holds
(i.e., there are no messages en route that can lead to a better solution).

Mattern's time algorithm is a variation of this basic method which allows checking the termination 
condition in only one wave.
Each work message (i.e., containing search states to be processed) has a time stamp,
which can be implemented as a clock counter maintained locally by each processor.
Every time a new termination check is started, the 
initiating processor increments its clock counter and sends a \emph{control} message to another processor,
starting a chain of control messages that will visit all processors and return to the first one.
When receiving a control message, a processor updates its clock counter $C$ to 
$\max (C, T)$, where $T$ is the maximum
clock value among processors visited so far. %
If a processor contains a received message $m$ with a time stamp $t_m \geq T$,
then the termination check fails.
Obviously, if, at the end of the chain of messages, the accumulated sent and received counters
differ, then the termination check fails as well.

In hash based work distribution, the choice of the hash
function is essential for achieving uniform distribution of the keys,
which results in effective load balancing.  Our implementation of HDA*
uses the Zobrist function~\cite{Zobrist:70} to map a SAS+ state
representation \cite{Backstrom:95} to a hash key.  The Zobrist
function is commonly used in the game tree search community to detect
duplicate states.  The Zobrist hash value is computed by XOR'ing
predefined random numbers associated with the components of a
state. When a new state is generated, the hash value of the new state
can be computed incrementally from the hash value of the parent state by incrementally XOR'ing the 
state components that differ.
The Zobrist function was previously used in domain-independent planning in MacroFF
\cite{macroff05}.
It is possible for two different states to have the same hash key, although the probability of such a collision is extremely low with 64-bit keys.
In MacroFF, as well as an earlier version of HDA* \cite{ICAPS09}, duplicate checking in the open/list was performed by checking if the hash key of a state was present in the open/closed list, so there was a non-zero (albeit tiny) probability of a false positive duplicate check result. In this paper, our HDA* implementations perform duplicate checks by
comparing the actual states. Although this is slightly slower than
comparing only the hash key, duplicate checks are guaranteed to be
correct.

 \newif\ifnewspecs 
\newspecstrue
\newcommand{\pea}{a 2.33GHZ dual quad-core Xeon L5410 machine with $6 \times 2$ MB L2 cache and 32GB memory}
\newcommand{\multicore}{{\tt Multicore} }
\newcommand{\tsubametwo}{{\tt HPC2} }
\newcommand{\tsubameone}{{\tt HPC1} }
\newcommand{\funcluster}{{\tt Commodity} }
\newcommand{\citemachinetable}{ (see Table~\ref{machine-descriptions} for machine specs)}

\section{Scalability of HDA*}
\label{sec:results-scaling}

We experimentally evaluated 
HDA* on top of a domain-independent planner and an application-specific, 24-puzzle solver.
Our hardware environments, including a single, multicore machine ({\tt Multicore}), a commodity cluster ({\tt Commodity}),
and two high-performance clusters (HPC clusters) with up to 2400 processors ({\tt HPC1}, {\tt HPC2}), are shown in Table \ref{machine-descriptions}.
In all of our experiments, HDA* is implemented in \textrm C++, compiled with {g++} 
and parallelized using the MPI message passing library.

\ifnewspecs
\begin{table}[htbp]
\begin{center}
\begin{tabular}{|l||r|r|r|l|}
 \hline
 System &    node description &   \# cores & RAM & Interconnection    \\
 Name   &                     &  per node  & per node & between nodes \\
\hline
\hline
\multicore   & 2.33GHz 2x 4-core & 8 &  32GB & n/a \\
             & Xeon L5410             &   &  &      \\
\hline
\funcluster    &  2.33GHz 2x 4-core & 8 & 16GB & 1Gbps(x2)\\
            & Xeon L5410         &   &  &    Ethernet  \\
\hline
\tsubameone & 2.4GHz 8x 2-core & 16 & 32GB & 20Gb Infiniband\\
          & AMD Opteron & & & \\
\hline
\tsubametwo & 2.93GHz 2x 6-core & 12 & 54GB & QDR \\
            & Xeon X5670        &    &      & Infinibandx2 \\
          &  & & & (80 Gbps)\\
 \hline
\end{tabular}
\caption{Parallel machines used in experiments}
\label{machine-descriptions}
\end{center}
\end{table}

\fi %

We first describe experimental results 
for domain-independent planning.
We parallelized the sequential optimal version of 
the Fast Downward planner,
enhanced with the so-called LFPA heuristic, 
which is based on explicit (merge-and-shrink) state abstraction~\cite{Helmert:07}.
All the reported results are obtained with the abstraction size set to 1,000.
Preliminary experiments with the abstraction size set to 5,000 did not change
the results qualitatively.
As benchmark problems, we use classical planning instances 
from past planning competitions.
We selected instances that are hard or unsolvable for the sequential optimal version of
Fast Downward.

HDA*, like other asynchronous parallel search algorithms,
behaves nondeterministically, resulting in some differences in search behavior between
identical invocations of the algorithms. However, on the runs where we
collected multiple data points, we did not observe significant differences between runs of HDA*. 
Therefore, due to the enormous resource requirements of a large-scale
experimental study,\footnote{In addition to usage charges for the clusters, there are issues of resource contention because the clusters are shared among hundreds of users.}
the results shown are for single runs.

\subsection{Asynchronous vs Synchronous Communications: Experiments on a Single, Multicore Machine}
\label{sec:multicore-results}
\begin{table*}[htb]
\begin{center}
\begin{scriptsize}

\begin{tabular}{|r||r||r|r|r|r||r|r|r|r||r|@{}r@{}|}
 \hline
  & 1 core & \multicolumn{4}{|c||}{4 cores} & \multicolumn{4}{|c||}{8 cores} & \multicolumn{2}{|c|}{}\\
  \hline
  & A* & \multicolumn{2}{|c|}{\tiny HDA*} & \multicolumn{2}{|c||}{\tiny PRA*}  & \multicolumn{2}{|c|}{\tiny HDA*} & \multicolumn{2}{|c||}{\tiny PRA*} &  Abst & Opt \\
             & time   & spd & eff & spd & eff &  spd & eff & spd & eff & time & len \\
\hline
{\bf Average} & {\bf 971.72} & {\bf 2.71} & {\bf 0.68} & {\bf 2.56} & {\bf 0.64} & {\bf 4.98} & {\bf 0.62} & {\bf 4.45} & {\bf 0.56} & {\bf 2.76} & \\
\hline
        Depot10 &    99.82 & 2.51 &  0.63 & 2.45 &  0.61 & 4.61 &  0.58 & 4.11 &  0.51 &  1.96 &  24 \\
\hline
        Depot13 &  1561.83 & 2.66 &  0.67 & 2.57 &  0.64 & 4.78 &  0.60 & 4.44 &  0.56 &  4.16 &  25 \\
\hline
     Driverlog8 &   102.55 & 2.59 &  0.65 & 2.14 &  0.54 & 4.90 &  0.61 & 3.87 &  0.48 &  0.14 &  22 \\
\hline
      Freecell5 &   137.01 & 2.99 &  0.75 & 2.99 &  0.75 & 5.87 &  0.73 & 5.83 &  0.73 &  7.75 &  30 \\
\hline
      Freecell7 &  2261.67 & 3.12 &  0.78 & 3.06 &  0.77 & 5.88 &  0.74 & 5.85 &  0.73 &  9.60 &  41 \\
\hline
        Rover12 &   923.23 & 2.73 &  0.68 & 2.46 &  0.61 & 5.03 &  0.63 & 4.43 &  0.55 &  0.10 &  19 \\
\hline
     Satellite6 &   104.83 & 2.25 &  0.56 & 2.06 &  0.51 & 4.30 &  0.54 & 3.61 &  0.45 &  0.09 &  20 \\
\hline
      ZenoTrav9 &   157.98 & 2.57 &  0.64 & 2.29 &  0.57 & 4.61 &  0.58 & 3.66 &  0.46 &  0.19 &  21 \\
\hline
     ZenoTrav11 &   424.68 & 2.42 &  0.60 & 2.10 &  0.52 & 4.40 &  0.55 & 3.60 &  0.45 &  0.22 &  14 \\
\hline
    PipesNoTk14 &   248.77 & 3.91 &  0.98 & 3.65 &  0.91 & 5.50 &  0.69 & 4.74 &  0.59 &  1.68 &  30 \\
\hline
    PipesNoTk24 &  1046.94 & 2.97 &  0.74 & 2.72 &  0.68 & 5.37 &  0.67 & 4.82 &  0.60 &  5.99 &  24 \\
\hline
       Pegsol27 &   178.71 & 2.97 &  0.74 & 2.89 &  0.72 & 5.87 &  0.73 & 5.09 &  0.64 &  1.11 &  28 \\
\hline
       Pegsol28 &   773.36 & 2.98 &  0.75 & 2.94 &  0.73 & 5.87 &  0.73 & 4.92 &  0.61 &  0.65 &  35 \\
\hline
      Airport17 &   322.21 & 3.54 &  0.89 & 3.52 &  0.88 & 6.62 &  0.83 & 6.77 &  0.85 & 13.09 &  88 \\
\hline
       Gripper8 &   304.82 & 2.57 &  0.64 & 2.29 &  0.57 & 4.41 &  0.55 & 3.63 &  0.45 &  0.28 &  53 \\
\hline
       Gripper9 &  1710.39 & 2.63 &  0.66 & 2.31 &  0.58 & 4.51 &  0.56 & 4.05 &  0.51 &  0.36 &  59 \\
\hline
       Mystery6 &   315.21 & 3.08 &  0.77 & 3.01 &  0.75 & 5.64 &  0.70 & 5.39 &  0.67 & 17.51 &  11 \\
\hline
         Truck5 &   365.38 & 2.17 &  0.54 & 2.32 &  0.58 & 4.23 &  0.53 & 4.00 &  0.50 &  0.26 &  25 \\
\hline
         Truck6 &  3597.24 & 2.41 &  0.60 & 2.23 &  0.56 & 4.53 &  0.57 & 4.11 &  0.51 &  0.30 &  30 \\
\hline
         Truck8 &  2194.38 & 2.39 &  0.60 & 2.23 &  0.56 & 4.35 &  0.54 & 3.82 &  0.48 &  0.22 &  25 \\
\hline
      Sokoban19 &   157.82 & 2.90 &  0.72 & 2.57 &  0.64 & 5.61 &  0.70 & 4.76 &  0.60 &  1.59 & 164 \\
\hline
      Sokoban22 &   428.91 & 3.07 &  0.77 & 3.03 &  0.76 & 5.99 &  0.75 & 5.27 &  0.66 &  1.53 & 172 \\
\hline
     Blocks10-2 &   327.03 & 2.83 &  0.71 & 2.39 &  0.60 & 5.51 &  0.69 & 4.24 &  0.53 &  1.04 &  34 \\
\hline
Logist00-7-1 &  1235.26 & 2.40 &  0.60 & 2.16 &  0.54 & 4.43 &  0.55 & 3.75 &  0.47 &  0.07 &  44 \\
\hline
Logist00-9-1 &  2082.76 & 2.46 &  0.61 & 2.45 &  0.61 & 4.44 &  0.55 & 4.32 &  0.54 &  0.13 &  30 \\
\hline
   Miconic12-2 &  2308.03 & 2.11 &  0.53 & 2.19 &  0.55 & 3.76 &  0.47 & 3.46 &  0.43 &  0.08 &  40 \\
\hline
   Miconic12-4 &  2463.13 & 2.15 &  0.54 & 2.18 &  0.54 & 3.87 &  0.48 & 3.55 &  0.44 &  0.08 &  41 \\
\hline
       Mprime30 &  1374.27 & 2.52 &  0.63 & 2.49 &  0.62 & 4.58 &  0.57 & 4.64 &  0.58 &  7.18 &   9 \\
\hline
\end{tabular}

\end{scriptsize}
  \caption{
{\small Comparison of sequential A*, HDA* and PRA* (without node retraction)
on 1, 4, and 8 cores on the  \multicore machine.
Runtimes (in seconds), speedup (spd), efficiency (eff), abstraction heuristic initialization times (not included in runtimes in previous columns), and optimal plan length are shown.}
}
\label{tab:1machine}
\end{center}
\end{table*}

First, we evaluate HDA* on a single, multicore machine (presented in Table~\ref{machine-descriptions})
in order to investigate the impact of asynchronous vs synchronous communications in parallel A*.
We compare HDA* with sequential A* and a shared-memory implementation
of 
Parallel Retracting A* (PRA*)  \cite{Evett:1995} 
on a single, multicore machine.
As described in Section \ref{sec:background}, PRA* uses the same
hash-based work distribution strategy as HDA*, but uses synchronous communications. 
As in Burns et al.'s experiments \cite{Burns:09}, our PRA* implementation does not include the node retraction scheme
because the main goal of our experiments is to show the impact of eliminating synchronization overhead from PRA*. 

Our HDA* implementation is the same MPI-based implementation used in our larger-scale, distributed memory experiments described below. It executes a separate OS process for each thread of execution, and we rely on the message passing functions in MPI for asynchronous communications.
While it may be possible to develop a more efficient implementation of HDA* specifically for shared memory machines (e.g., using kernel threads and shared memory constructs), 
we wanted to investigate the scalability of the same HDA* implementation on both shared and distributed memory machines.

Low level optimizations were implemented to make both the HDA* and PRA* as fast as possible.
In addition to locks available in the Boost C++ library, %
we also incorporated spin locks based on the ``xchgl'' assembly operation %
in order to speed up 
PRA*.
Each algorithm used the full 32 gigabytes of RAM available on the machine at hand.
That is, $n$-core HDA* spawns $n$ processes, each using 32/$n$
gigabytes of RAM, sequential A* used the full 32GB available, and the
multithreaded PRA* algorithm shared 32GB of RAM among all of
the threads.

Table \ref{tab:1machine} shows the speedup of HDA* and PRA* for 4 cores and 8 cores.
We show only instances that can be solved by the serial planner with 32GB RAM available.
In addition to runtimes for the sequential A* algorithm, 
the speedup and the parallel \emph{efficiency}
are shown for HDA* and PRA*. Efficiency 
is defined as $S/P$, 
where $S$ is the speedup over a serial run and $P$ is the number of cores.
As shown in Table \ref{tab:1machine}, HDA* clearly outperforms 
PRA*.
With 4 cores, the speedup of HDA* ranges from 2.11 to 3.91,
and the efficiency ranges from 0.53 to 0.98.
With 8 cores, the speedup of HDA* ranges from 3.76 to 6.62,
and the efficiency ranges from 0.47 to 0.83.
These results demonstrate the benefit of asynchronous message passing over a synchronous implementation of hash-based work distribution.

\subsection{Planning Experiments on a HPC Cluster}
\label{sec:planning-cluster-results}

Next, we investigate the scaling behavior of HDA* on 
the \tsubametwo  cluster, (see machine specs in Table~\ref{machine-descriptions}).
We used 1-200 nodes in our experiments (i.e., 1-2400 cores).

\begin{table}[bthp]
\begin{center}
\begin{scriptsize}
\begin{tabular}{|@{}l@{}||c|c|c|c|c|c|c|c||@{}c@{}|@{}r@{}|
}
\hline
           & 12  & 24  & 60  & 144             & 300             & 600  & 1200  & 2400             & Abst  & Opt      \\
\hline
Freecell7 & 175.87 &   95.27 &   38.31 &   18.11 &   10.69 &   10.96 &   18.11 &   68.40 & 4.73 & 41 \\
\hline
Logistics00-7-1 & 115.71 &   56.41 &   23.88 &   10.78 &    6.50 &    6.80 &   18.76 &   48.34 & 0.09 & 44 \\
\hline
Logistics00-9-1 & 175.72 &   91.48 &   38.43 &   17.13 &    9.34 &    7.73 &   18.64 &   61.43 & 0.13 & 30 \\
\hline
Mprime30 & 123.85 &   71.82 &   32.07 &   14.92 &    8.11 &    6.55 &    8.03 &   28.38 & 4.10 & 9 \\
\hline
Pegsol.p28 &  52.89 &   31.59 &   12.90 &    6.38 &    4.41 &   14.73 &   21.90 &   57.18 & 0.52 & 35 \\
\hline
PipeNoTank24 &  89.33 &   49.88 &   19.99 &    9.05 &    5.55 &    5.69 &   19.42 &   65.30 & 3.07 & 24 \\
\hline
Probfcell-4-1 & 112.07 &   60.52 &   25.64 &   11.82 &    6.75 &    5.90 &    9.18 &   31.57 & 1.62 & 19 \\
\hline
Rover6 & 445.17 &  249.76 &  101.71 &   44.46 &   23.27 &   14.19 &   11.57 &   18.76 & 0.12 & 36 \\
\hline
Rover12 &  42.42 &   26.36 &   10.67 &    5.10 &    3.61 &    4.14 &    8.07 &   26.03 & 0.12 & 19 \\
\hline
Sokoban.p24 & 136.43 &   79.95 &   34.00 &   17.64 &   26.79 &   27.43 &   51.88 & -        & 0.69 & 205 \\
\hline
Sokoban.p28 & 171.97 &   93.00 &   39.95 &   19.47 &   25.13 &   25.75 &   49.70 & -        & 0.62 & 135 \\
\hline
ZenoTravel11 &  38.49 &   21.84 &    8.28 &    4.33 &    3.15 &    4.77 &   13.61 &   28.61 & 0.20 & 14 \\
\hline
Freecell6 &-         &  386.55 &  163.18 &   70.85 &   37.14 &   22.03 &   20.85 &   52.59 & 4.36 & 34 \\
\hline
Pegsol.p29 &-         &  146.08 &   61.49 &   27.78 &   15.08 &   11.18 &   14.11 &   52.84 & 6.45 & 37 \\
\hline
PipesTank10 &-         &  689.17 &  292.89 &  126.54 &   62.68 &   33.81 &   22.01 &   33.78 & 4.25 & 19 \\
\hline
Satellite7 &-         & 1339.41 &  415.04 &  147.22 &   63.39 &   33.40 &   24.50 &   37.55 & 0.19 & 21 \\
\hline
Sokoban.p25 &-         &  108.73 &   44.00 &   22.23 &   19.46 &   26.40 &   47.44 & -        & 1.06 & 155 \\
\hline
DriverLog13 &-         & -         &  163.65 &   66.60 &   33.82 &   20.31 &   18.54 &  130.18 & 0.25 & 26 \\
\hline
Logistics00-8-1 &-         & -         &  222.63 &   95.03 &   91.91 &   26.18 &   27.19 &   46.58 & 0.09 & 44 \\
\hline
Logistics00-9-0 &-         & -         &  210.95 &   94.24 &   45.07 &   25.67 &   22.48 &   65.00 & 0.13 & 36 \\
\hline
Mprime24 &-         & -         &   60.60 &   33.53 &   16.97 &   10.72 &   10.78 &   23.86 & 25.31 & 8 \\
\hline
Pegsol.p30 &-         & -         &  168.18 &   75.79 &   39.51 &   22.39 &   20.54 &   47.44 & 1.30 & 48 \\
\hline
PipeNoTank18 &-         & -         &  131.98 &   58.09 &   29.46 &   18.77 &   15.93 &  141.66 & 2.28 & 30 \\
\hline
PipesTank9 &-         & -         &  418.51 &  180.63 &   89.31 &   47.85 &   29.22 &   34.26 & 5.43 & 18 \\
\hline
Probfcell-5-3 &-         & -         &  733.73 &  315.30 &  156.25 &   81.34 &   47.80 &   47.86 & 2.48 & 24 \\
\hline
Sokoban.p26 &-         & -         &   95.44 &   47.36 &   33.79 &   34.26 &   70.25 & -        & 1.04 & 135 \\
\hline
Sokoban.p27 &-         & -         &  167.96 &   79.19 &   48.15 &   33.52 &   37.98 & -        & 1.54 & 87 \\
\hline
Depot16 &-         & -         & -         &  461.64 &  226.00 &  116.86 &   64.08 &   58.78 & 3.97 & 25 \\
\hline
Freecell11 &-         & -         & -         &  304.71 &  151.55 &   81.74 &   54.63 &   99.66 & 6.10 & 56 \\
\hline
Mprime15 &-         & -         & -         &  128.58 &   61.32 &   32.74 &   21.52 &   26.42 & 21.04 & 6 \\
\hline
PipeNoTank20 &-         & -         & -         &  109.74 &   51.77 &   34.32 &   20.59 &  120.82 & 3.54 & 28 \\
\hline
PipeNoTank32 &-         & -         & -         &  121.60 &   61.85 &   32.90 &   22.81 &   92.47 & 3.68 & 30 \\
\hline
PipesTank14 &-         & -         & -         &  126.94 &   62.53 &   34.69 &   23.94 &   37.51 & 6.00 & 38 \\
\hline
PipesTank22 &-         & -         & -         &  183.47 &   90.66 &   50.08 &   31.02 &   50.57 & 8.30 & 30 \\
\hline
Probfcell-5-1 &-         & -         & -         &  659.28 &  323.29 &  165.22 &   89.94 &   62.69 & 3.10 & 24 \\
\hline
Probfcell-5-2 &-         & -         & -         &  375.00 &  185.41 &   95.11 &   56.24 &   89.90 & 2.80 & 23 \\
\hline
Probfcell-5-4 &-         & -         & -         &  352.93 &  174.56 &   90.15 &   51.50 &   46.54 & 3.76 & 23 \\
\hline
Probfcell-5-5 &-         & -         & -         &  546.76 &  268.79 &  137.54 &   76.12 &   59.21 & 2.36 & 25 \\
\hline
ZenoTravel12 &-         & -         & -         &  291.60 &  138.88 &   72.40 &   48.09 &   57.17 & 0.24 & 21 \\
\hline
Freecell9 &-         & -         & -         & -         &  334.06 &  172.59 &   96.56 &   93.50 & 6.49 & 43 \\
\hline
PipeNoTank33 &-         & -         & -         & -         &  185.34 &   97.02 &   54.70 &   55.72 & 5.04 & 32 \\
\hline
PipeNoTank35 &-         & -         & -         & -         &  205.55 &  105.50 &   58.83 &   53.94 & 7.63 & 22 \\
\hline
Freecell12 &-         & -         & -         & -         & -         &  268.28 &  148.74 &  117.21 & 5.32 & 47 \\
\hline
PipeNoTank25 &-         & -         & -         & -         & -         &  266.94 &  142.34 &  101.08 & 3.57 & 32 \\
\hline
PipeNoTank27 &-         & -         & -         & -         & -         &  319.07 &  170.78 &  103.62 & 5.14 & 26 \\

\hline
\end{tabular}
\caption{Planning runtimes on the \tsubametwo cluster\citemachinetable.}
\label{tab:tsubame2-planning-runtimes}
\end{scriptsize}
\end{center}
\end{table}

As in Table~\ref{tab:1machine}, the times shown in Table~\ref{tab:tsubame2-planning-runtimes}
include the time for the search
algorithm execution, and {\em exclude} the time required to compute
the abstraction table for the LFPA heuristic, since this phase of Fast
Downward+LFPA has not been parallelized yet
and therefore requires
the same amount of time to run regardless of the number of
cores. 
For example, the IPC-6 Pegsol-30
instance, which requires 168.18 seconds with 60 cores, was solved in
20.54 seconds with 1200 cores, plus 1.30 seconds for the abstraction
table generation.
A value of  ``-'' for the runtimes in Table \ref{tab:tsubame2-planning-runtimes} indicates a
failure, i.e., the planner terminated because one of the nodes ran out of memory.
For example, the Freecell-12 instance was first solved using 600 cores.

Let $t_k$ be the runtime to solve a problem using $k$ processors.
Standard metrics for evaluating parallel performance on $n$ processors include speedup and efficiency, defined as 
as $S_n = t_1/t_n$, and $E_n = S_n/n$, respectively.
These standard metrics have limited applicability for parallel A* because sequential runtimes cannot be measured for hard problems.
On hard problems that truly require large-scale parallel A*, sequential A* exhausts RAM and
terminates before finding a solution. Since sequential runtime $t_1$ cannot be measured, $S_n$ and $E_n$ cannot be computed.
While we could use only benchmarks which can be solved using a single processor (as we did for our multicore experiment in Section \ref{sec:multicore-results}),
this would restrict the benchmark set to problems that can be solved by A* using only the RAM available on 1 processing node.
Suppose there is 4.5GB RAM per core, as is the case with our \tsubametwo cluster.
A single-threaded A* algorithm that generates 50,000 new states per second, using 50 bytes per state, will exhaust 4.5GB within 30 minutes. 
In fact, because the state size for difficult planning instances is usually much larger than 50 bytes,
serial A* exhausts 4.5GB memory much more quickly. 
Problems that require only a few minutes to solve using sequential search are poor benchmarks for large-scale parallel search, because such problems can be solved in a few seconds using 1000 processors. 

Mahapatra and Dutt also noted that sequential runtimes were not
available for their scalability experiments \cite{MahapatraD97}. They evaluated their algorithms by computing  {\em approximate speedups}, defined as follows.
Let $p_{min}$ be the minimum number of processors that solved the
problem. They make the assumption that the speedup with $p_{min}$ processors is
$p_{min}$ (i.e., they assume that linear speedup can be obtained up to $p_{min}$ processors). Based on this, they estimate that sequential runtime is $p_{min} \times t_{p_{min}}$,
and speedups for $p>p_{min}$ processors can be computed by computing the ratio of parallel runtime to this estimate of
sequential runtime. 

Mahapatra and Dutt argue that this is a conservative estimate of speedup because overheads increase as the number of processors increases, so the assumption that $S_p=p_{min}$ is a lower bound on the the actual speedup for $p \geq p_{min}$.
However, there are some issues with this approach:
(1) such estimates of ``speedup'' are not actual speedup values,
(2) $p_{min}$ can be different for different problem instances,  and
(3) when $p_{min}$ is large,  e.g., if a problem is first solved using 600 cores, including these estimated speedups significantly biases computations of average speedup to seem artificially high -- in other words, assuming that linear speedups are obtainable up to $p_{min}$ processors is unrealistic when $p_{min}$ is large.

Thus, we take a different approach to measuring the scalability of parallel search, based on the performance of $n$ processors relative to the performance on $p_{min}$.
Let $p_{min}$ be the smallest number of cores that solves a problem, and $t_{min}$ be the wall-clock runtime for $p_{min}$ cores.
The {\em relative speedup} is defined as $S_n = t_{min}/t_n$, and
the {\em relative efficiency} is defined as $E_n = S_n / (n/p_{min})$. 
Relative efficiency has been used by other researchers, c.f., Niewiadomski et al. \cite{Niewiadomski06} (who called it ``speedup efficiency'').

A second issue related to measuring performance on a cluster of
multicore machines is the baseline configuration of a single processing node.
As shown later in Section \ref{sec:scale-nodes}, the number of cores used per processing node has a significant effect on performance. Our main goal is to investigate the scalability of HDA* while fully utilizing all available processors, so in our scalability experiments, we use all cores on a processor, e.g., on the \tsubametwo cluster,
we use the 12 cores per core, allocating 4.5GB RAM per core.

Table \ref{tab:tsubame2-planning-runtimes} shows the wall-clock runtimes of HDA* on a set of 45 IPC benchmark problems.
The results are grouped according to  the smallest number of processors that solved the problem.
As mentioned earlier, a ``-'' indicates a failure, i.e., the planner terminated because one of the cores ran out of memory.
For example, the Depot16 instance was first solved using 144 cores.
Sokoban.p24-Sokoban.p27 failed with 2400 cores, even though they could be solved with fewer cores.

\begin{figure}
\begin{center}
\includegraphics[width=.6\textwidth]{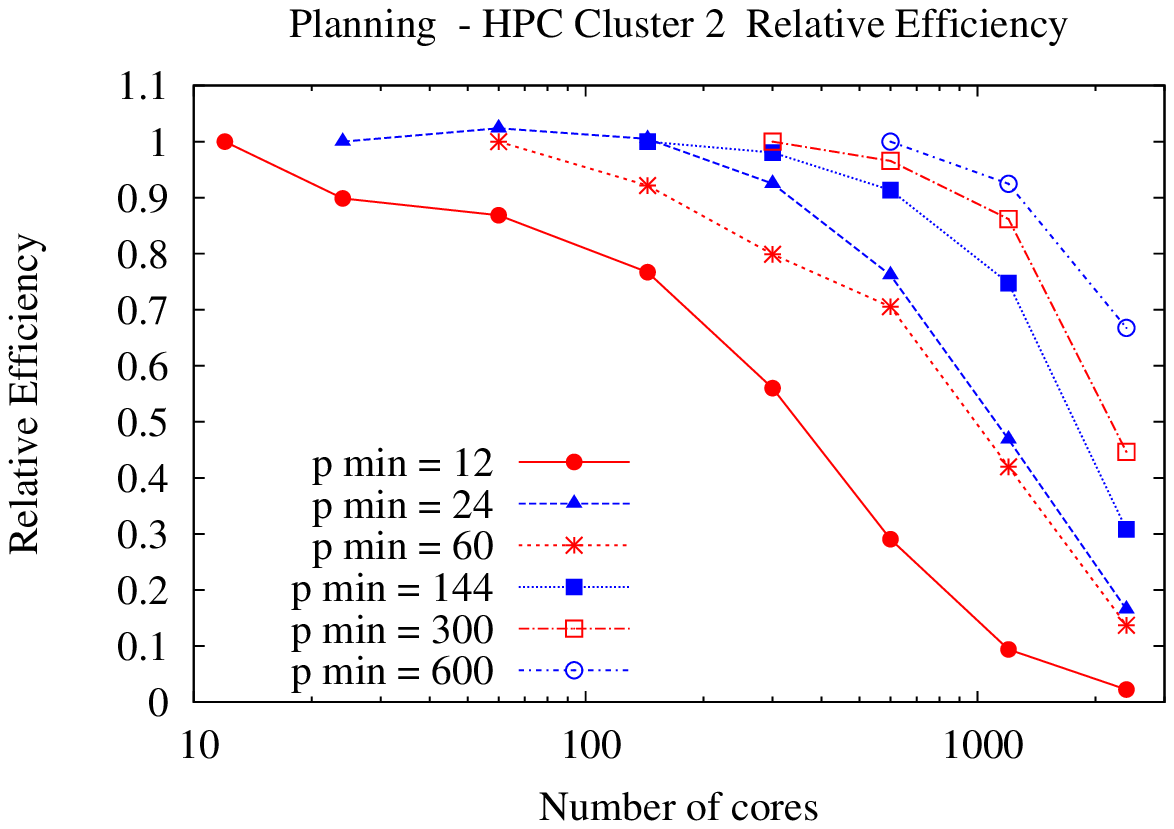}
\end{center}
\caption{Planning relative efficiency on the \tsubametwo cluster\citemachinetable.
}
\label{fig:tsubame2-planning-relative-efficiency}
\end{figure}

Figure \ref{fig:tsubame2-planning-relative-efficiency} shows that the relative efficiency (see above) of HDA* on planning generally scales well for $2p_{min}$ and $4p_{min}$ cores, even when $p_{min}=600$ cores. Even for $8p_{min}$ cores, the relative efficiency is over 0.5 for $p_{min} \in \{12,24, 60, 144, 300\}$.
However, as $p/p_{min}$ increases, the relative efficiency degrades. For $p_{min}=12$, when $p=1200$ ($p/p_{min}=100$), relative efficiency is 9.25\%, and for $p=2400$ cores, the relative efficiency is a mere 2.2\%.

In other words, HDA* scales reasonably efficiently for up to 4-8 times $p_{min}$, the minimum number of processors that can solve the problem.
However, as the number of processors is further increased, there is a point of diminishing returns, and eventually, adding more processors can result in {\em longer} runtimes.
An extreme case of this scaling limitation can be seen on the Sokoban benchmark problems (Sokoban.p24-p.28), which were solvable with up to 1200 cores (with diminishing returns), but terminated due to memory exhaustion at some processor on 2400 cores.
On the other hand, it is important to note that even for very large values of $p_{min}$ (e.g., 600 cores), HDA* continues to scale very efficiently for $4p_{min}$ (2400) cores.

\begin{figure}
\begin{center}
\includegraphics[width=.6\textwidth]{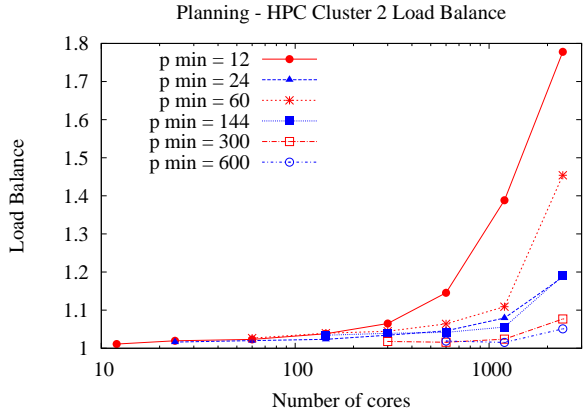}
\end{center}
\caption{Planning load balance on the \tsubametwo cluster.}
\label{fig:tsubame2-planning-load-balance}
\end{figure}

\subsubsection{Load Balance}
\label{sec:load-balance-planning}

A common metric for measuring how evenly the work is 
distributed among the cores is the {\em load balance}, defined as
the ratio of the maximal number of states expanded by a core and the average number of states expanded by each core.
As shown in Figure \ref{fig:tsubame2-planning-load-balance}, HDA* achieves good load balance when $p/p_{min} \leq 8$.
While load balance tends to degrade as the number of processors
increases, it is important to note that load imbalance does not seem to be simply caused
by using a large, absolute number of processors.
For instance, 
on all 6 problems where $p_{min} \geq 600$, the load balance for 2400 cores is less than 1.10.

One possible reason for load imbalance %
may be ``hotspots''
 --  frequently generated duplicate nodes mapped to a small number of cores by the hash function.
This is caused by {\em transpositions} in the search space,
which are states that can be reached through different paths.
In HDA*, if a processor receives a state $s$ which is already in the closed list but the $g$-value
of $s$ is smaller than that in the closed list, $s$ must be enqueued in the open list.
However, the heuristic value of $s$ is not recomputed in saving $s$ to the open list, because the value 
is already saved in the closed list.
For example, in solving PipesNoTk24 with 2400 cores,
more than 70\% of generated states are duplicates. 
A processor involved in a hotspot receives about 377\% more
duplicate states than the processor receiving duplicate states least frequently,
although we observe that they receive similar amounts of work.
As a result, the numbers of calls for the heuristic function are different (about 144\%)
between these processors. 
Thus, a processor which executes fewer heuristic evaluations 
(relative to other processors receiving a comparable number of states) has a higher state expansion rate than the other processors, resulting in load imbalance.

\subsubsection{Search Overhead}
\label{so-planning}

The {\em search overhead}, which indicates the extra states explored by parallel
search, is defined as:
{\small 
\[
SO = 100 \times (\frac{\mbox{number of states expanded by parallel search}}
{\mbox{number of states expanded by baseline search}} - 1).
\]
}
This is the percentage of extra node expansions performed compared to a 
given baseline algorithm or hardware configuration.
For a direct measurement of the search overhead for some given instance
our baseline algorithm (configuration) will be HDA* on $p_{min} \in \{12,24,60,144,300,600,1200,2400\}$ cores, the minimal number of cores that solves that instance.
This has the advantage that, by definition, the baseline configuration will always succeed,
allowing a direct measurement of the search overhead.
As we show in this section, it is also possible to analyze the search overhead compared to
serial A* even in cases where serial A* fails.
Such comparisons can show how effective HDA* is in terms of wasted search effort.

We start by identifying possible causes of the search overhead 
in the case when
the baseline algorithm is serial A*. Let $c^*$ be the
cost of an optimal solution. Serial A* with a consistent heuristic in use
expands all states with $f<c^*$, and some of the states with $f=c^*$. No states with $f>c^*$
get expanded. No state is expanded more than once. In contrast, parallel
variants of A*, including HDA*, can re-expand states with $f \leq c^*$ which are not re-expanded by A*,
can expand states
with $f>c^*$, and can expand a different number of states with $f=c^*$ than serial
A*. Possible causes include incomplete
local knowledge  of open lists at other processors, and nondeterministic travel time and arrival order
of messages containing states.

Parallel A* usually expands more nodes than serial A*.
If we consider only nodes with $f<c^*$ or $f>c^*$,
the best that parallel A* can possibly achieve is matching the number of
expansions performed by serial A* with a consistent heuristic.
Thus, while it is possible, in principle, for parallel A* to expand fewer nodes than serial A* with a consistent heuristic, this is only possible if parallel A* expands fewer nodes with $f=c^*$ than serial A*.

\begin{figure}
\begin{center}
\includegraphics[width=.48\textwidth]{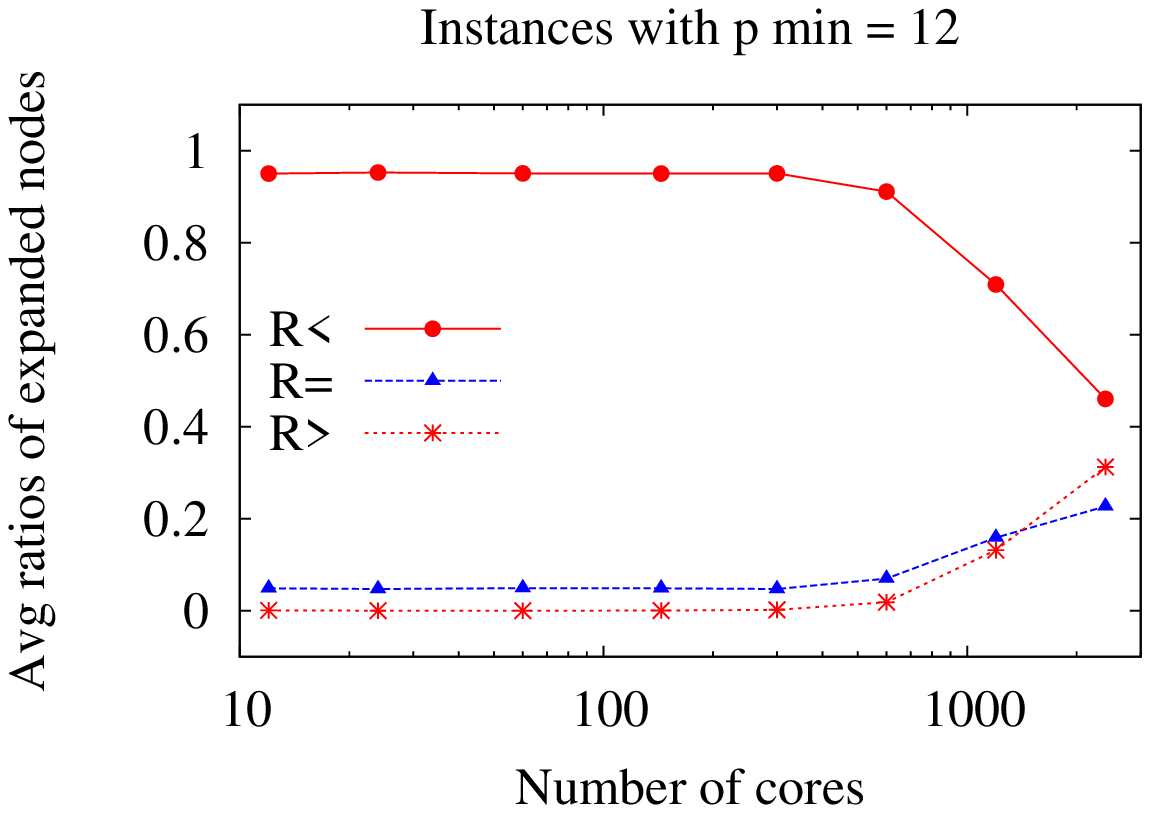}
\includegraphics[width=.48\textwidth]{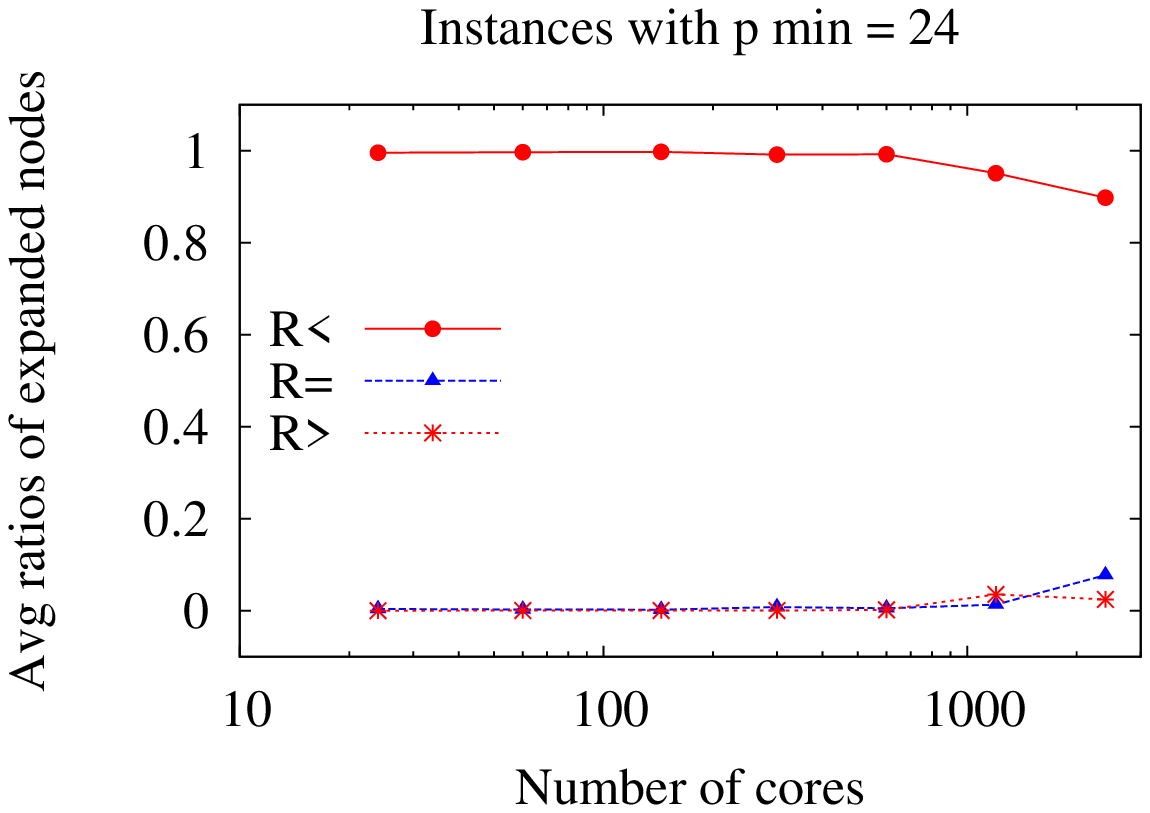}
\includegraphics[width=.48\textwidth]{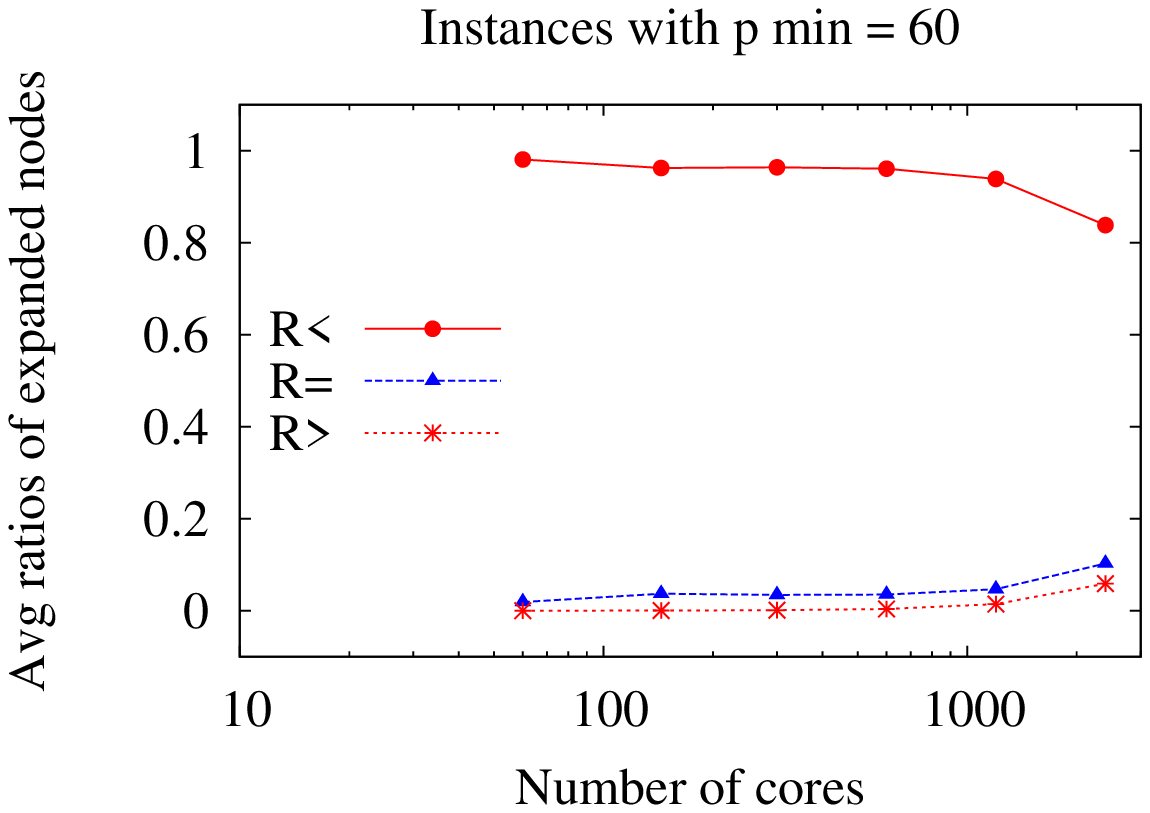}
\includegraphics[width=.48\textwidth]{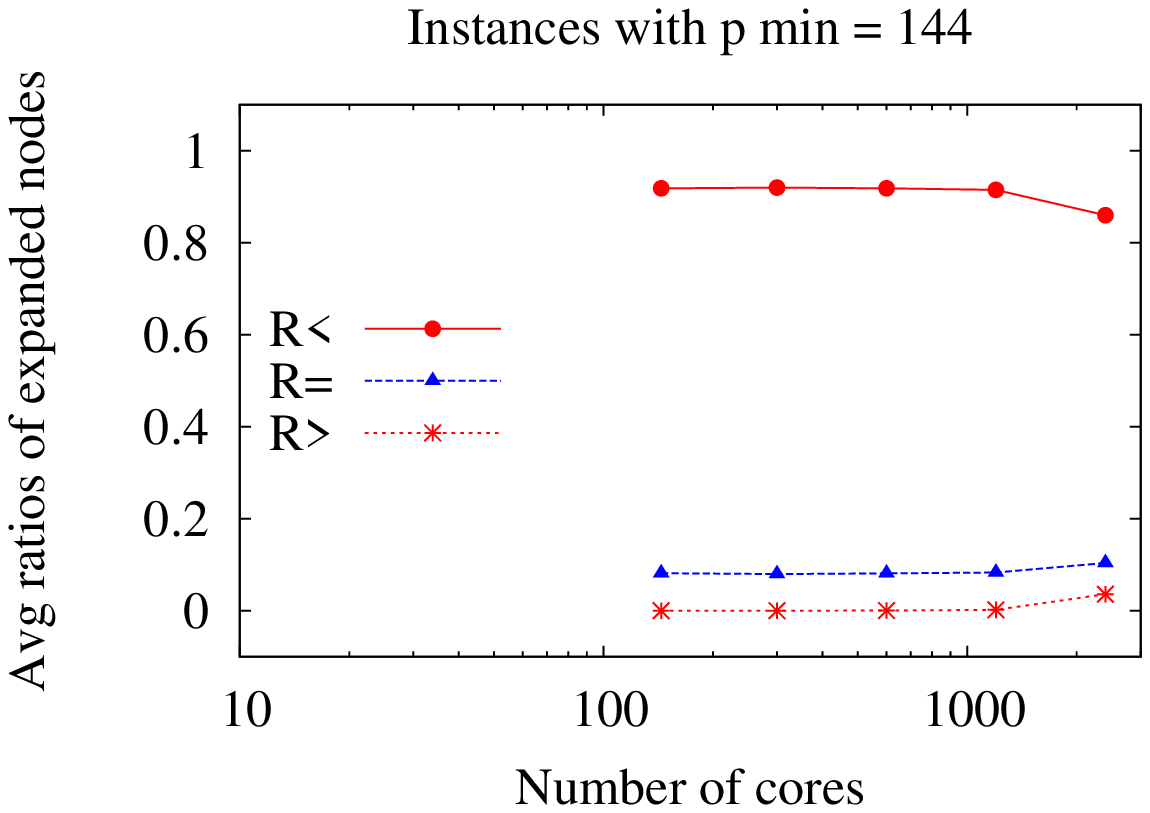}
\includegraphics[width=.48\textwidth]{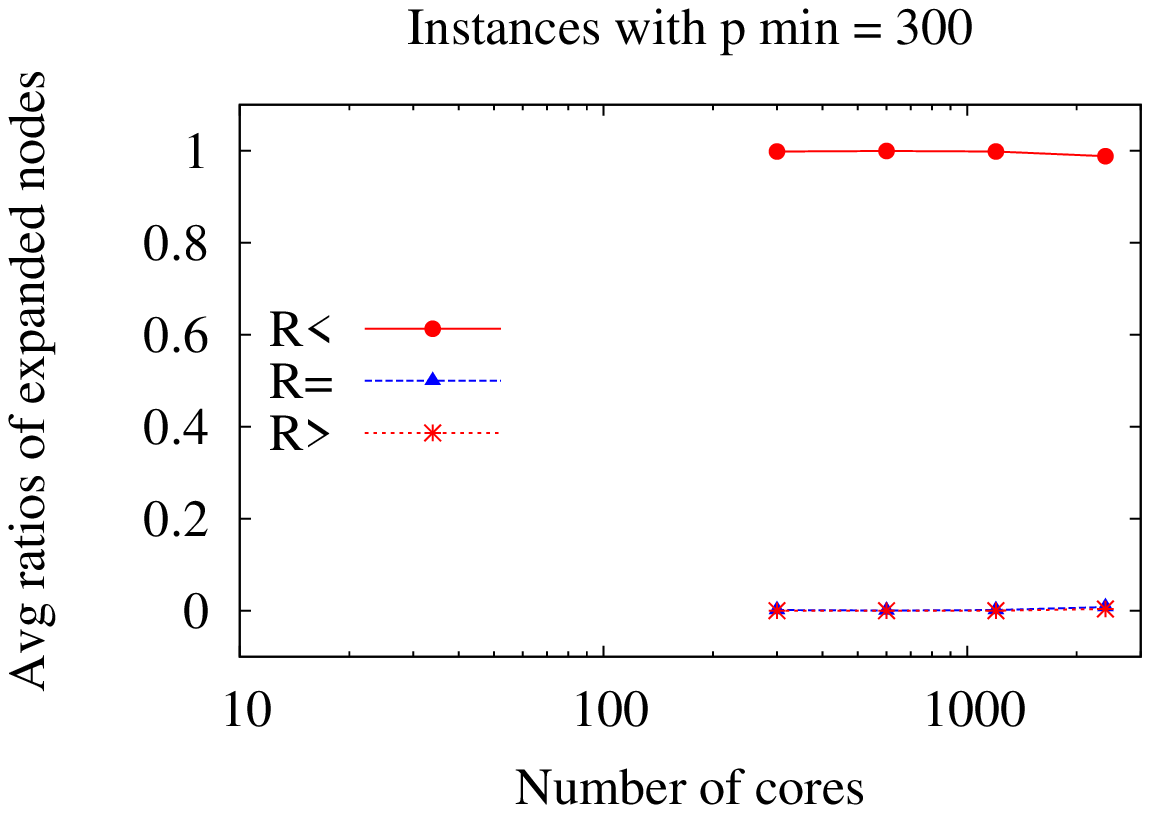}
\includegraphics[width=.48\textwidth]{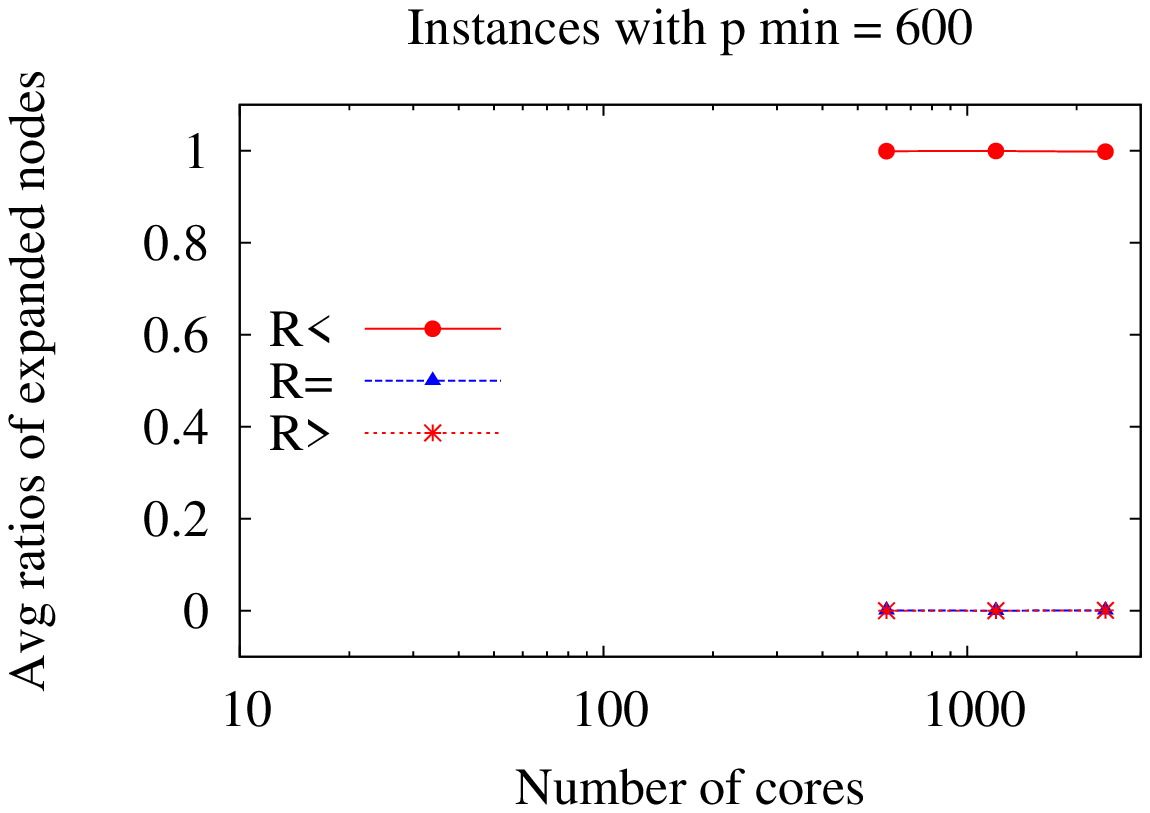}
\end{center}
\caption{Average values for $R_<$, $R_=$ and $R_>$, on the \tsubametwo cluster.
}
\label{fig:tsubame2-f-vs-opt-ratio}
\end{figure}

\begin{figure}
\begin{center}
\includegraphics[width=.48\textwidth]{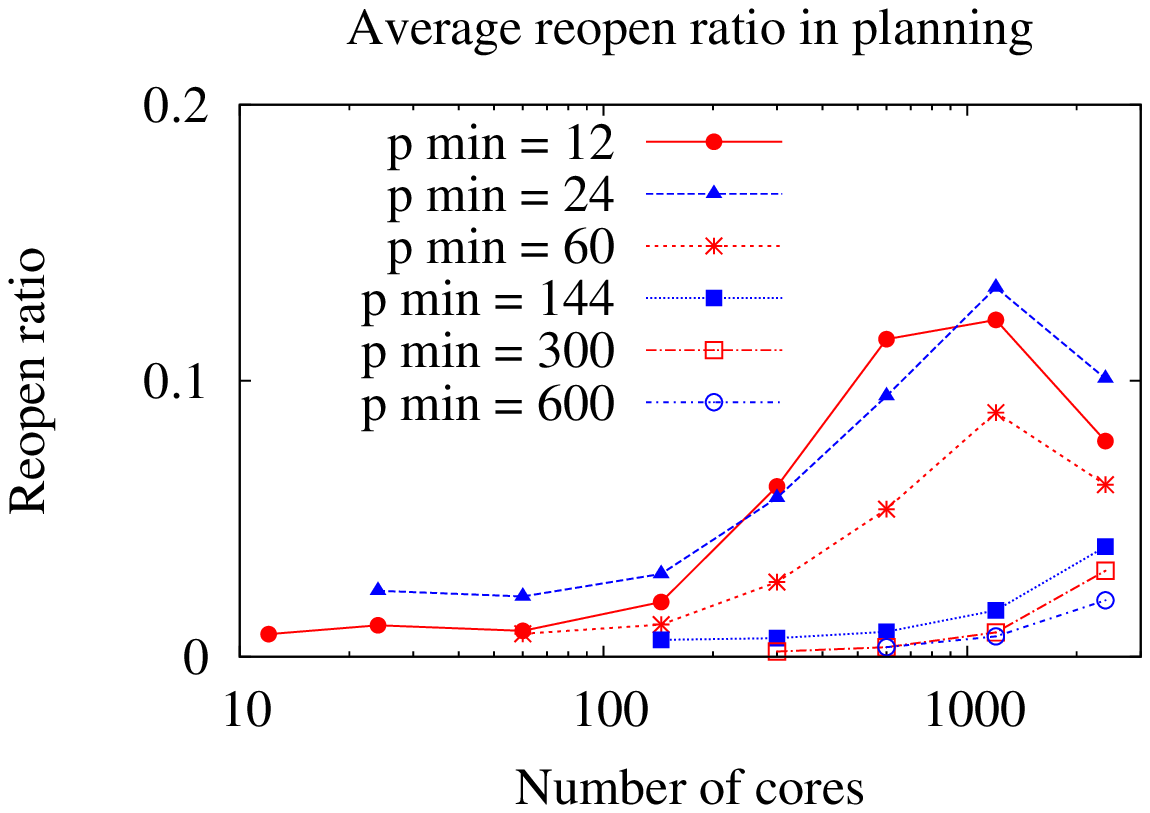}
\includegraphics[width=.48\textwidth]{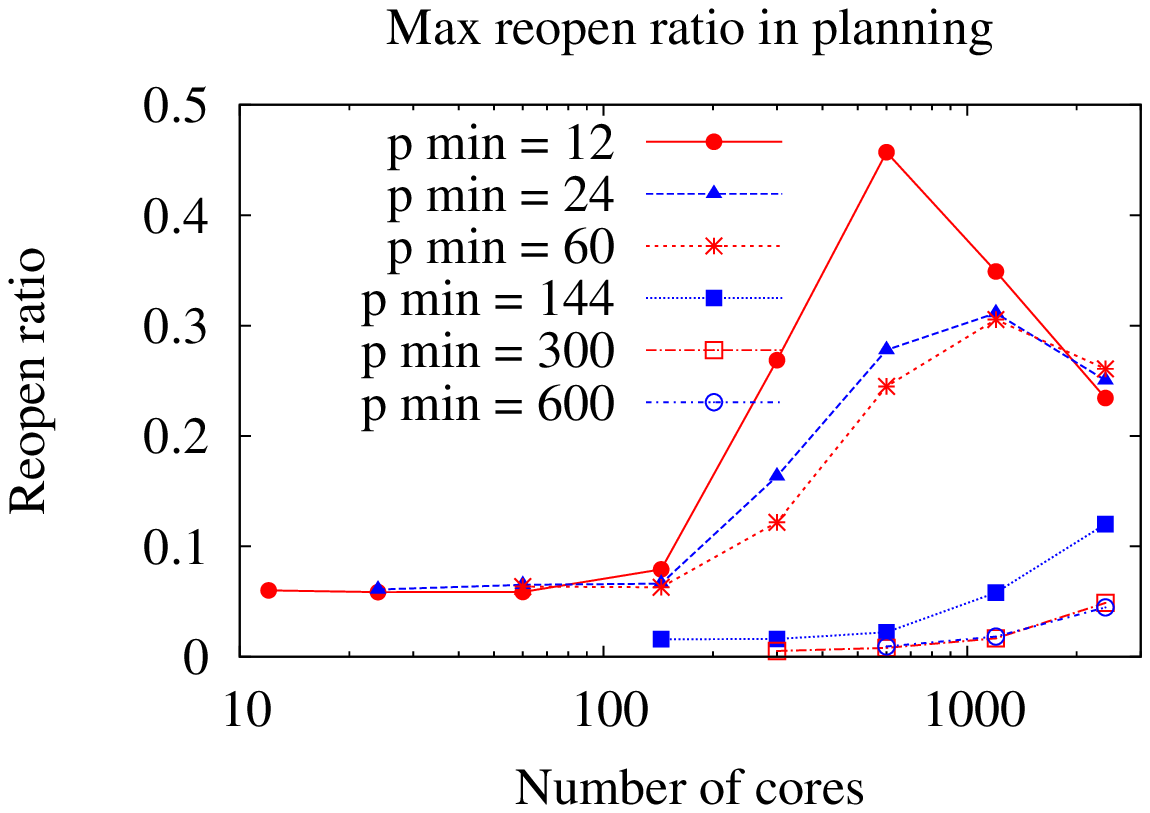}
\end{center}
\caption{Ratio of node re-expansions $R_r$, in planning, on the \tsubametwo cluster. 
Left: average values computed across all instances with the same $p_{min}$.
Right: maximum values.
}
\label{fig:tsubame2-reopen-ratio}
\end{figure}

Define $R_{<}$ as the fraction of expanded nodes with $f<c^*$.
$R_{=}$ and $R_{>}$ are defined similarly. 
Let $R_r$ be the ratio of node re-expansions, which is the total number of re-expansions
divided by the total number of expansions.
Figure~\ref{fig:tsubame2-f-vs-opt-ratio}
summarizes  $R_{<}$, $R_{=}$ and $R_{>}$ 
for domain-independent planning, according to $p_{min}$.
$R_r$, the ratio of re-expansions, is shown in Figure~\ref{fig:tsubame2-reopen-ratio}.

When serial A* fails to solve an instance, 
a direct measurement of the search overhead relative to serial A*
cannot be performed. 
Fortunately, in such cases, our ``$R_*$ metrics'' ($R_<$, $R_=$, $R_>$ and $R_r$) 
can allow us to
make inferences about the search overhead.
The key observation is that serial A* expands \emph{all} nodes with $f<c^*$.
Therefore, if %
HDA* exhibits low values for 
$R_=$, $R_>$ and $R_r$, on some run, we can conclude that the search overhead is low in that run.
In other words, in such cases, there is little wasted search effort
introduced by distributing the search.
{\em This appears to be the case for almost all instances, when using $p_{min}$ cores.}
A noticeable exception
is Mprime15 ($p_{min}=144$),
where $R_< = 6\%$ and $R_= = 94\%$.\footnote{This is because the optimal solution cost for Mprime15 is only 6 and 
there are many states with tied $f$ values.}
As $p$, the number of cores in use for a given instance,
grows increasingly larger than $p_{min}$, the $R_*$ metrics
maintain very good values for a while.
Then a degradation can eventually be observed,
for those instances where $p_{min} \leq 144$
(Figure~\ref{fig:tsubame2-f-vs-opt-ratio}).

To explain the good $R_{>}$ values of HDA* run on $p_{min}$ cores
and other core configurations as well (cf. Figure~\ref{fig:tsubame2-f-vs-opt-ratio}),
we start by recalling that serial A* with a consistent heuristic will expand
all nodes with $f=v$ before expanding any node with $f>v$.
Such a strict monotonicity cannot be guaranteed for HDA*.
However, we believe that HDA* can expand states almost monotonically, %
especially when the number of nodes with $f<c^*$
is large enough to make the instance challenging to a $p_{min}$-core configuration.
More specifically, consider an instance where the number of nodes with $f<c^*$ is large, and the number of nodes
with $f=c^*$ that end up being expanded by serial A* is very small in comparison.
Serial A*
will expand all nodes with $f<c^*$, after which it will start expanding
nodes with $f=c^*$.
Likewise, we hypothesize that HDA* could expand a majority of the nodes with $f<c^*$,
before starting expanding nodes with $f\geq c^*$.
If an optimal solution is found soon after starting expanding nodes with $f\geq c^*$,
the extra work caused by expanding nodes with $f>c^*$ would be a small fraction
of the total search effort of HDA*.
This informal explanation is consistent with the cases of low $R_>$ values observed among our data.

All benchmark problems in this paper have unit costs
for all transitions between states. We believe that this contributes
to the good (low) re-expansion rate $R_r$, when using $p_{min}$
cores and $p>p_{min}$ cores as well, unless $p$ gets much larger
than $p_{min}$ (see Figure~\ref{fig:tsubame2-reopen-ratio}).
Kobayashi et al.~\cite{KobayashiKW2011} have observed that
the re-expansion rate increases in domains with non-unit transition costs.

\begin{figure}
\begin{center}
\includegraphics[width=.48\textwidth]{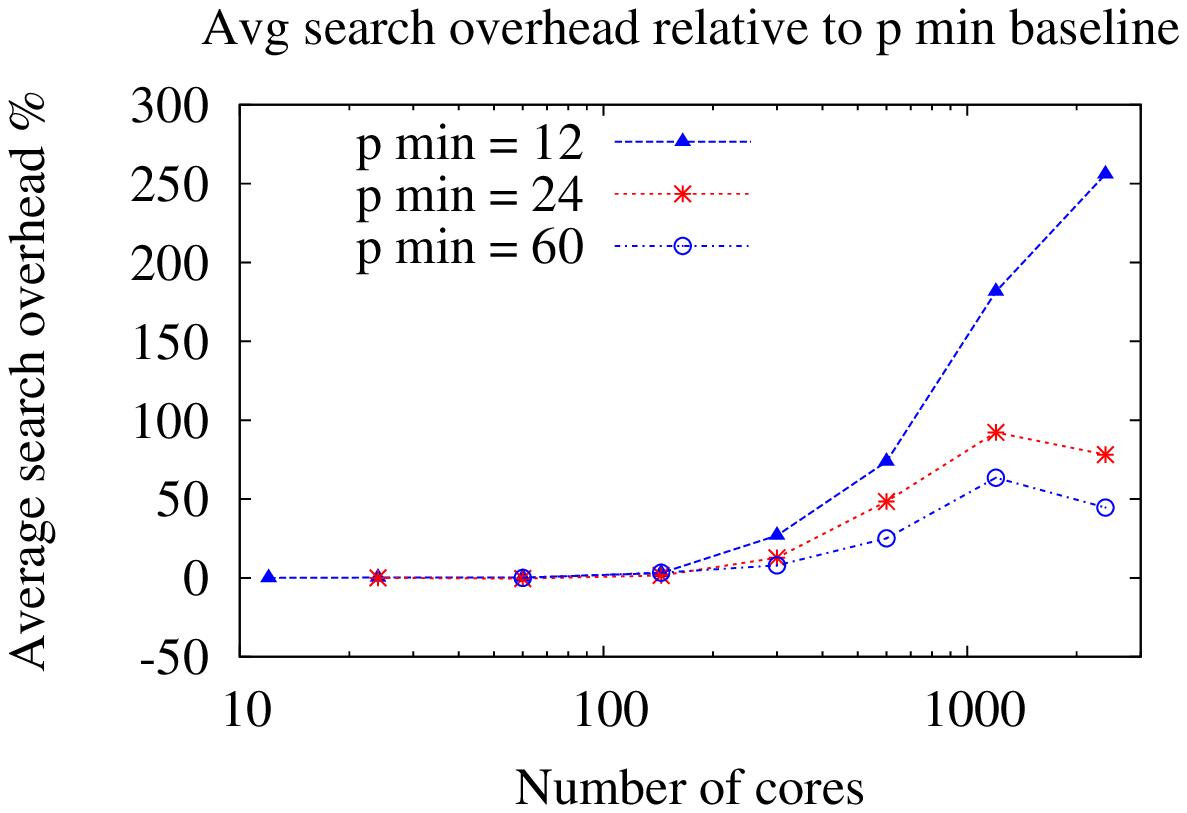}
\includegraphics[width=.48\textwidth]{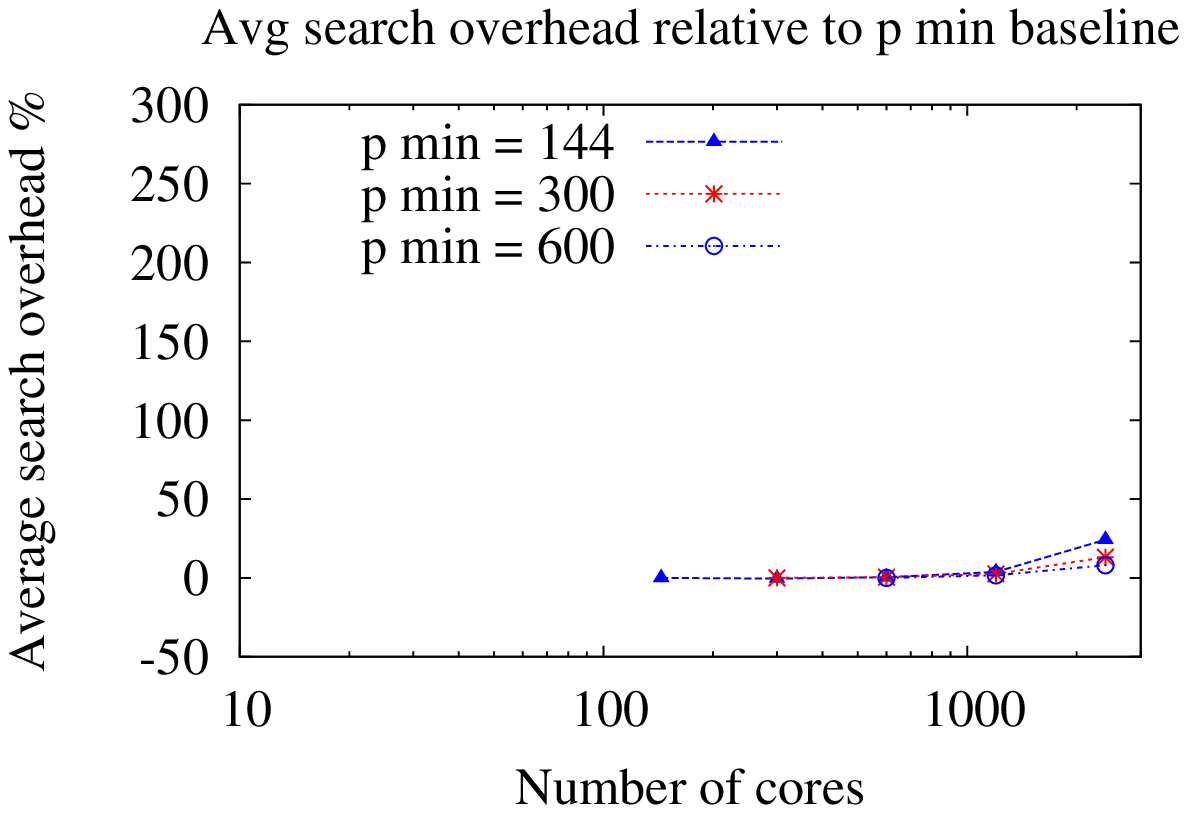}
\end{center}
\caption{Search overhead relative to $p_{min}$ cores, on the \tsubametwo cluster.
We plot data averaged over all instances with the same $p_{min}$.}
\label{fig:tsubame2-so-planning}
\end{figure}

Next we focus on the search overhead when the baseline algorithm is 
HDA* using $p_{min}$ cores 
(as defined above, $p_{min}$ depends on the instance).
Figure~\ref{fig:tsubame2-so-planning} summarizes search overhead data for domain-independent planning,
showing average values over all instances with the same $p_{min}$.

Once again, as a general tendency, 
as we go further away from the baseline $p_{min}$,
the search overhead stays stable for a while,
after which it grows (Figure~\ref{fig:tsubame2-so-planning}).
Thus, some of the largest search overheads 
are seen when the difference $p$ vs $p_{min}$ is the largest (2400 vs 12).
However, %
search overhead does not appear to be simply
caused by using a large number of processors.
For example, on 13 out of 45 problems, the search overhead on 2400 cores is less than 10\%.
This is despite the fact that $p_{min}$ is in these cases significantly lower than 2400,
varying from 24 cores (PipesTank10) to 600 cores (PipeNoTank25, PipeNoTank27).
There seem to be some problem domains which are highly
prone to search overhead. 
Very large search overheads compared to other domains are seen in Sokoban
problems (Sokoban.p24, Sokoban.p25, Sokoban.p26, Sokoban.p27, Sokoban.p28). 
The $R_*$ data available indicates that re-expansions are a major cause of this behavior.
In addition, on 1200 cores, the $R_<$ metric also degrades for a few Sokoban instances.
As the number of processors increased from 600 to 1200, 
the search overhead on Sokoban.p24, Sokoban.p26, and Sokoban.p27, more than doubles.
The growth in search overhead appears to be accelerating as $p/p_{min}$
increases.
This large, rapidly growing search overhead
explains the failure to solve the Sokoban problems with 2400 cores. As
we increase from 1200 to 2400 cores, the amount of search overhead
added is greater than the additional aggregate RAM capacity, so HDA*
fails because some processor runs out of RAM. 

Note that there are some instances where search overhead is negative relative
to $p_{min}$. For example, Mprime30 has a negative overhead for $p \in \{60, 144, 300\}$,
relative to $p_{min} = 12$.
This suggests some wasted search effort in the baseline configuration,
which then gets corrected for a larger $p$.
Indeed, 
the $R_<$ value is $62.2\%$
for Mprime30 solved on $p_{min}=12$ cores,
whereas most other instances, in all considered domains, have much better $R_<$ values (i.e., close to $100\%$) on their
corresponding $p_{min}$ configuration. 
Since Mprime has much shorter solutions than the other instances, HDA* tends to expand many more states
with the $f$-value identical to the optimal length.  

In summary, our analysis shows that
HDA* is scalable, in the sense that it can use
a large number of CPUs to solve difficult instances (which fail on a small number of CPU cores) quite efficiently, with a reasonably low search overhead.
At the same time, for easy instances that can be solved with fewer CPUs,
the tendency is the following: As the number of processors increases from $p_{min}$,
the search overhead stays low for a while, after which a degradation is observed.
One possible approach to alleviate this is using a solving strategy able to dynamically change the
hardware resources (e.g., number of CPUs)~\cite{FKB12}.
Mechanisms for keeping the search overhead low when using many more processors than $p_{min}$
is an interesting topic for future work.

\subsubsection{Node Expansion Rate}
\label{node expansion rate}

The node expansion rate helps evaluate other parallel overhead, besides the search overhead.
As the number of processors increases,
the communication overhead can reduce node expansion rate.
Part of the communication overhead is alleviated by the fact that 
searching and travelling overlap to a great extent.
Said in simple words, some states are being expanded while some
other states are travelling to their owner processor.

Since each core sends generated successors to their home processors in HDA*,
the total number of messages exchanged among processors increases
with a larger number of cores.
This is exacerbated by the fact that more cores can end up generating more states (search overhead).
Therefore, another cause of  slower node expansion rates may be message processing overhead --
even with asynchronous communications, 
HDA* must deal with a larger number of messages as the number of cores increases.

\begin{figure}
\begin{center}
\includegraphics[width=.48\textwidth]{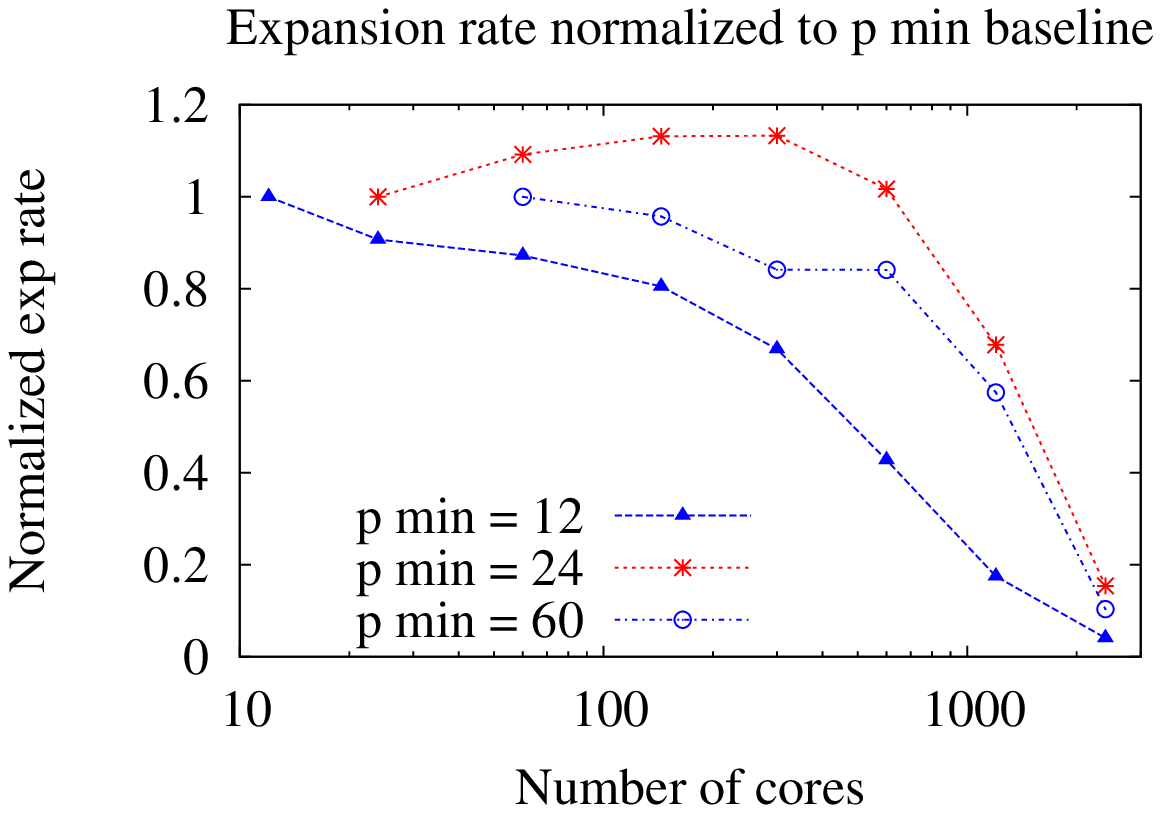}
\includegraphics[width=.48\textwidth]{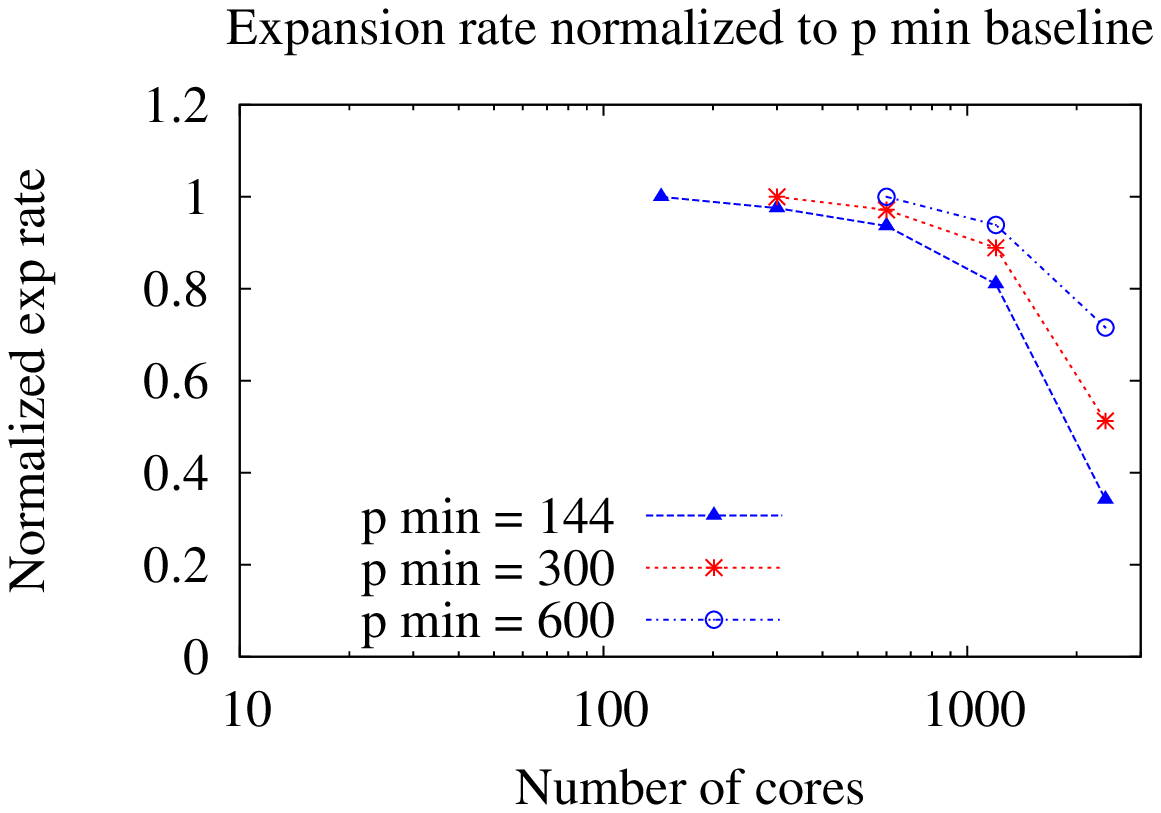}
\end{center}
\caption{Node expansion rate relative to the $p_{min}$ cores baseline, on the \tsubametwo cluster.}
\label{fig:tsubame2-comm-ovhd}
\end{figure}

Figure~\ref{fig:tsubame2-comm-ovhd} plots the expansion rate,
measured for planning instances on the \tsubametwo cluster.
Each data point in the figure is computed as follows.
Let $N^\iota_p$ be the total number of expansions for an instance $\iota$ solved on $p$
cores. Let $T^\iota_p$, the {\em total} search time to solve instance $\iota$, be the (parallel) wall-clock time multiplied by $p$. 
Then, the node expansion rate is $N^\iota_p/T^\iota_p$. 
Let $A(p, p_{min})$ be the expansion rate, when using $p$ cores,
averaged over all instances $\iota$ with a given $p_{min}$ value.
Finally, $R(p, p_{min}) = A(p, p_{min})/A(p_{min}, p_{min})$ normalizes these average
values relatively to the baseline configuration with $p_{min}$ cores.
Figure~\ref{fig:tsubame2-comm-ovhd} plots $R(p, p_{min})$. 
If $R(p,p_{min})$ is close to 1, this indicates that the communication overhead is low.
Values smaller than 1 indicate a degradation of the expansion rate, 
typically resulting in a degradation of the parallel search efficiency,
unless the increase in time spent per node is compensated by a corresponding reduction in the 
total number of node expansions.

Figure~\ref{fig:tsubame2-comm-ovhd} shows that the expansion rate remains almost constant relative to $p_{min}$ cores,
unless the difference $p - p_{min}$ grows too large. Note that due to the logarithmic scale of the X-axis (\# of cores), the expansion rate is more stable than it might initially appear in Figure~\ref{fig:tsubame2-comm-ovhd}.
Despite the degradation of node expansion rate at each processor, the aggregate node expansion rate across all processors continues to increase as the number of processors increases.
For example, when $p = 2400$ and $p_{min} = 12$, the expansion rate degrades by a factor
of about 22, according to Figure~\ref{fig:tsubame2-comm-ovhd}.
On the other hand, the number of available CPUs increases by a factor or $2400/12 = 200$, so 
the net gain in aggregate node expansion rate is $200/22 = 9.1$ times.
Similarly, there are 
substantial overall gains in the aggregate expansion rate, for all $p$ and $p_{min}$ values considered.

\subsubsection{Termination Detection}
Termination detection is not a performance bottleneck, as it succeeds quickly after proving solution optimality.
When an optimal (but not proven optimal yet) solution is found, its cost $c^*$ is broadcast, such that nodes with $f \geq c^*$ will not be expanded from now on. Thus, the only nodes expanded after the broadcast (if any) are nodes with $f<c^*$, which must be expanded anyway to prove optimality (serial A* expands them as well). Therefore, by the time $c^*$ has been broadcast \emph{and} all nodes with $f<c^*$ have been expanded, state expansion and generation will stop, messages with states will not be sent around any longer, and the termination test will succeed.

\subsubsection {Scaling Behavior and the Number of Nodes (Machines) on a HPC Cluster}
\label{sec:scale-nodes}

So far, we have considered the scaling of HDA* as the number of cores was increased.
However, the scaling behavior of HDA* on a cluster is affected by
factors other than the number of cores.  Another factor is the cost of
communication between nodes. Communications between cores in
the same node are done via a shared memory bus, while communications
between cores residing on different nodes are performed over
a local area network (e.g., Infiniband or Ethernet).

To evaluate the impact of the communication delay, 
we ran the planner on the \tsubameone cluster, whose specs are shown in Table~\ref{machine-descriptions}.
We used 64 cores, where the cores were distributed evenly on 4-64 nodes (i.e., 1-16 cores per node).
The results are shown in the upper half of Table \ref{tab:64-core-dummy}.
If the communication delay was a significant factor, we would expect that, as the number of nodes increased, the runtime would increase.
Interestingly, Table \ref{tab:64-core-dummy} (upper half) shows that runtimes {\em decreased} on almost all of the problem instances as the number of nodes increased.

\begin{table}[ht]
\begin{center}
\begin{footnotesize}

\begin{tabular}{|c|c|c|c|c|c|c|c|c|c|c|c|}
\hline
\multicolumn{7}{|c|}{Normal execution of HDA*}\\
\hline
            & 64 cores & 64 cores & 64 cores & 64 cores & 64 cores & Opt \\
            & 4 nodes & 8 nodes  & 16 nodes & 32 nodes & 64 nodes & Len \\
\hline
{\bf Avg time}  & {\bf 117.56} & {\bf 110.88} & {\bf 93.31} & {\bf 92.28} & {\bf 87.70} & \\
\hline
Freecell7   & 66.85     & 68.61     & 64.29      & 59.24      & 59.39      &  41 \\
\hline
Satellite7  & 502.51    & 468.07    & 370.43     &  375.86    & 351.69     &  21 \\
\hline
ZenoTrav11  & 16.58     & 15.64     & 15.29      &  14.71     & 14.41      &  14 \\
\hline
PipesNoTk24 & 40.34     & 39.85     & 38.08      &  35.34     & 34.53      &  24 \\
\hline
Pegsol28    & 21.77     & 21.31     & 20.75      &  20.47     & 19.64      &  35 \\
\hline
Sokoban24   & 57.29     & 51.82     & 50.99      &  48.08     & 46.51      &  205 \\
\hline
\hline
\multicolumn{7}{|c|}{HDA*, with dummy processes on cores not used by HDA*}\\
\hline
{\bf Avg time} & {\bf 117.56} & {\bf 113.87} & {\bf n/a} & {\bf 111.31} & {\bf 113.88} & \\
\hline
Freecell7   & 66.85    & 70.59    & 72.13    & 77.49    & 76.15    &  41 \\
\hline
Satellite7  & 502.51   & 466.16   & 535.86   & 445.64   & 462.22   &  21 \\
\hline
ZenoTrav11  & 16.58    & 21.17    & 20.15    & 19.89    & 20.54    &  14 \\
\hline
PipesNoTk24 & 40.34    & 44.99    & 45.26    & 43.79    & 43.80    &  24 \\
\hline
Pegsol28    & 21.77    & 22.88    & n/a      & 23.23    & 23.44    &  35 \\
\hline
Sokoban24   & 57.29    & 57.45    & 56.07    & 57.92    & 57.16    &  205 \\
\hline
\end{tabular}
\caption{64-core scaling results with no dummy (upper half) and with dummy processes (bottom half).
Time and plan length are shown. 
Notice the increase in the average time caused by adding dummy processes.
Experiments were performed on the \tsubameone cluster.
}
\label{tab:64-core-dummy}
\end{footnotesize}
\end{center}
\end{table}

The explanation for this counterintuitive result is memory contention within each single node.
The limited bandwidth of the memory bus is saturated when many cores in the node 
simultaneously access memory.
We further investigated this hypothesis as follows.
First, cache contention was ruled out as a possible explanation because
the processors in the cluster are based on the  AMD Opteron architecture, where each core has a private L1 and L2 cache.
Then, we performed an experiment where, again, we used 4-64 nodes, but this time, on each node, on each core that was not used by HDA*, we executed
a dummy process, which makes no direct contribution towards solving the instance at hand.
Each dummy process is an invocation of HDA* which is forced to run sequentially on the instance, and 
therefore behaves similarly to the ``real'' HDA* process with respect to memory access patterns (and therefore contends for memory with the ``real'' HDA* process).

Table \ref{tab:64-core-dummy} indicates that
the overhead of local memory access congestion within a node is 
much larger than the overhead due to communication between nodes.
In the normal configurations without dummy processes (Table \ref{tab:64-core-dummy}, upper half), using fewer cores per node results in less local memory contention and more inter-node communication. Overall, the performance improves as the 64 cores are distributed among an increasing number of nodes.
On the other hand, in the configurations with dummy processes (Table
\ref{tab:64-core-dummy}, bottom half), local memory access within a node becomes
equally congested regardless of the number of nodes, so there is no
benefit to distributing the cores.
Once the local memory access
overhead is equalized for all configurations (4-64 nodes),
variations in performance (if any present) among configurations
could be explained by the overhead of inter-node communication.
We notice, however, that the differences are small in row ``Avg time'' in the
bottom half of Table \ref{tab:64-core-dummy}.
Even on a per-instance basis,
the performance degradation as the number of
nodes increases from 4 to 64 is within 20\%.
This shows that communications among nodes is not a critical bottleneck for HDA* on the \tsubameone cluster.

\subsection{Results on the 24-Puzzle on a HPC Cluster}
\label{24-puzzle results}

We evaluated HDA* on the 24-puzzle, using an application-specific solver based on IDA* code provided by Rich Korf,
which uses a disjoint pattern database heuristic \cite{KorfF02}.
We replaced the IDA* search strategy with A* and HDA*.
In the original code, each state has two redundant halves, trading memory for a faster state processing.
While this makes sense for IDA*, it is not the best trade-off for HDA*.
Thus, we reduce the state size to one half and compute the missing half on demand with a loop of
25 iterations per state.
Other parts of the program, including the disjoint pattern database heuristic, were unchanged.
As benchmark instances, we used the 50-instance set of 24-puzzles reported in \cite{KorfF02}, Table 2.
We excluded the time required to read pattern databases from the hard-disk.

\begin{figure}
\begin{center}
\includegraphics[width=.6\textwidth]{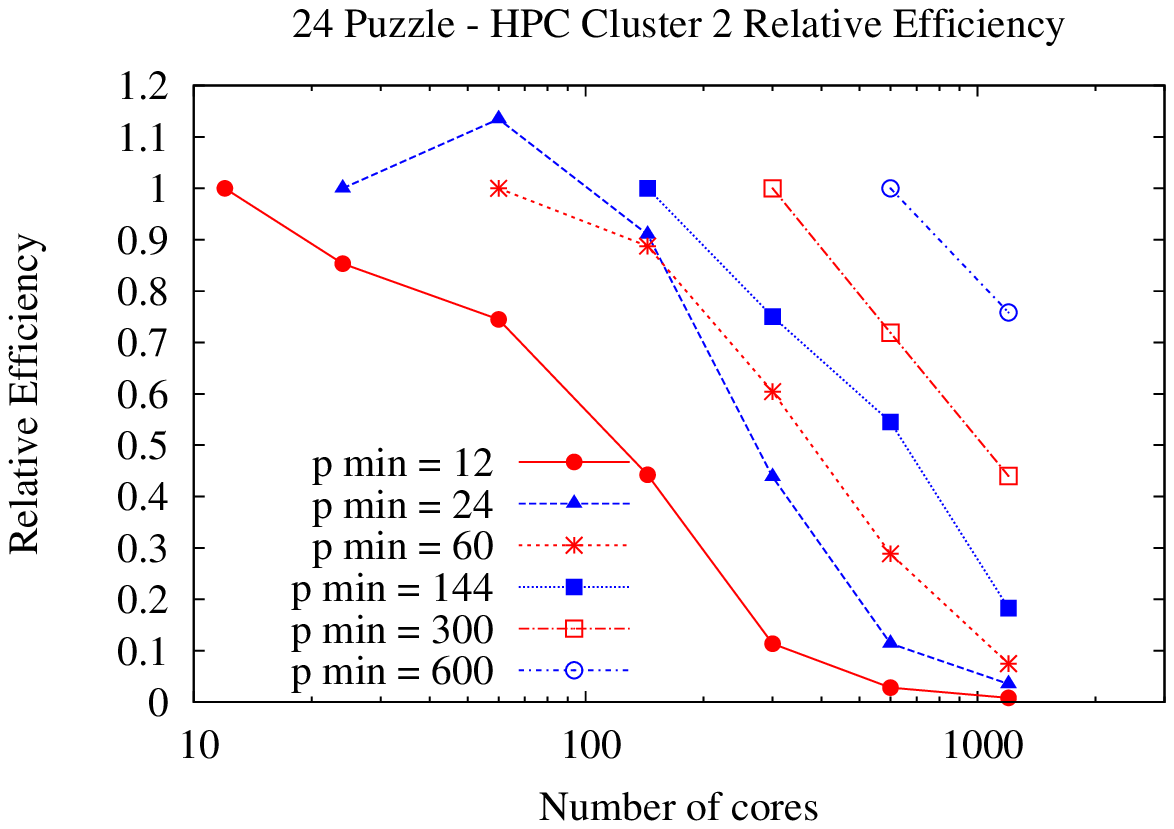}
\end{center}
\caption{24-puzzle relative efficiency on the \tsubametwo cluster. 
}
\label{fig:tsubame2-24puz-relative-efficiency}
\end{figure}

First, we investigated the scalability of HDA* by running it on the fifty 24-puzzle instances, using 1-1200 cores on 1-100 processing nodes.
As with domain-independent planning, we computed the efficiencies relative to the smallest number of cores $p_{min} \in \{12,24,60,144,300,600,1200 \}$ which solved the problem. These average relative efficiencies for each $p_{min}$ are shown in Figure \ref{fig:tsubame2-24puz-relative-efficiency}.
Compared to domain-independent planning (Figure \ref{fig:tsubame2-planning-relative-efficiency}), the relative efficiency of HDA* degrades much more rapidly on the 24-puzzle.
As previously mentioned in Section \ref{sec:introduction},
a key difference between domain-independent planning and the 24-puzzle is 
that individual states are processed much faster in the application specific 24-puzzle solver, and therefore parallel overheads have a greater weight in the total running time.

Table~\ref{tab:tsubame2-24-puzzle-scaling-by-mincores} provides summary data 
obtained on the \tsubametwo cluster\citemachinetable.
Instances are partitioned according to $p_{min}$.
For each partition, we report the runtime, the speedup, 
the load balance,
the search overhead relative to $p_{min}$ cores,
and the $R_*$ metrics (i.e., $R_{<}$, $R_{=}$, $R_{>}$, $R_{r}$)
related to the search overhead, introduced in Section~\ref{so-planning}.
All shown values are averages over instances with the same $p_{min}$.
The load balance is quite good, and is no worse than 1.17 (300 and 600 cores for $p_{min}=12$).
For $p_{min}$ processors, the search overhead is small.
 $R_r$ and $R_{>}$ are within 4\%, even for $p_{min}$ = 1200. 
$R_{=}$ varies from 8\% ($p_{min} \in \{144, 300\}$) to 24\% ($p_{min} = 24$).
As with our planning results, search overhead increases, as more processors are used and the difference between $p$ and $p_{min}$ increases.
Not surprisingly, the largest increase is seen when $p_{min}=12$ and $p=1200$ cores.
When an instance is easy enough to be solved with 12 cores, 1200 cores will do much redundant work, resulting in 
$R_{>}$ as high as 90\%.
The search overhead has a significant impact on the overall speedup reported in
Table~\ref{tab:tsubame2-24-puzzle-scaling-by-mincores}.
As $p$ grows larger and larger than $p_{min}$,
the speedup increases for a while, after which the search overhead dominates
and the speedup degrades.

The trends in search overhead for the 24-puzzle are similar to those observed for domain-independent planning.
On difficult instances (large $p_{min}$), HDA* can make an effective use of a large number of cores,
solving such instances with a reasonably low search overhead.
For instances that are sufficiently easy to be solved with relatively few cores (small $p_{min}$),
the performance (e.g., speedup) increases for a while, after which a degradation is observed.
A remarkable difference from the planning data is that the search overhead is significantly
higher in the 24-puzzle, and it has a sharper degradation rate as well.
\renewcommand{\arraystretch}{.85}
\begin{table}[htb]
\begin{center}
\begin{scriptsize}
\begin{tabular}{|c|c|c|c|c|c|c|c|c|}
\hline
 &    &  {p = \bf 12 }  &  {\bf 24 } & {\bf 60 } &  {\bf 144 } &   {\bf 300 }  &  {\bf 600 }  &  {\bf 1200 } \\
\hline
p min & Runtime  & 22.65& 14.17& 5.60& 3.86& 10.68& 15.71& 26.64\\
= 12 & Speedup  & 1.00& 1.71& 3.72& 5.31& 2.84& 1.40& 0.81\\
 & Load balance  & 1.05& 1.11& 1.07& 1.06& 1.17& 1.17& 1.14\\
 & Search overhead \% & 0.00& 13.66& 32.38& 257.81& 1955.27& 5099.60& 16977.93\\
 & $R_<$  & 0.87& 0.78& 0.75& 0.55& 0.19& 0.05& 0.02\\
 & $R_=$  & 0.13& 0.22& 0.24& 0.34& 0.47& 0.25& 0.08\\
 & $R_>$  & 0.00& 0.00& 0.01& 0.12& 0.34& 0.70& 0.90\\
 & $R_r$  & 0.01& 0.01& 0.01& 0.02& 0.04& 0.08& 0.07\\
\hline
p min & Runtime  & -& 48.33& 17.12& 8.87& 9.28& 16.97& 27.36\\
= 24 & Speedup  & -& 1.00& 2.84& 5.46& 5.48& 2.85& 1.76\\
 & Load balance  & -& 1.14& 1.09& 1.07& 1.12& 1.15& 1.12\\
 & Search overhead \%  & -& 0.00& -16.63& -7.97& 47.75& 438.86& 1599.57\\
 & $R_<$  & -& 0.76& 0.91& 0.83& 0.54& 0.16& 0.05\\
 & $R_=$  & -& 0.24& 0.09& 0.16& 0.40& 0.64& 0.26\\
 & $R_>$  & -& 0.00& 0.00& 0.01& 0.06& 0.20& 0.69\\
 & $R_r$  & -& 0.01& 0.01& 0.02& 0.03& 0.07& 0.06\\
\hline
p min & Runtime  & -& -& 50.09& 22.98& 16.21& 17.01& 33.09\\
= 60 & Speedup  & -& -& 1.00& 2.13& 3.02& 2.89& 1.50\\
 & Load balance  & -& -& 1.09& 1.07& 1.13& 1.14& 1.16\\
 & Search overhead \% & -& -& 0.00& 4.33& 52.52& 221.04& 965.01\\
 & $R_<$  & -& -& 0.90& 0.88& 0.68& 0.36& 0.11\\
 & $R_=$  & -& -& 0.10& 0.12& 0.26& 0.51& 0.53\\
 & $R_>$  & -& -& 0.00& 0.00& 0.05& 0.13& 0.36\\
 & $R_r$  & -& -& 0.01& 0.01& 0.03& 0.06& 0.06\\
\hline
p min & Runtime  & -& -& -& 53.51& 34.61& 23.46& 35.50\\
= 144 & Speedup  & -& -& -& 1.00& 1.56& 2.27& 1.52\\
 & Load balance  & -& -& -& 1.07& 1.09& 1.13& 1.13\\
 & Search overhead \% & -& -& -& 0.00& 14.55& 52.61& 338.69\\
 & $R_<$  & -& -& -& 0.92& 0.82& 0.65& 0.24\\
 & $R_=$  & -& -& -& 0.08& 0.16& 0.31& 0.66\\
 & $R_>$  & -& -& -& 0.00& 0.01& 0.05& 0.09\\
 & $R_r$  & -& -& -& 0.01& 0.03& 0.05& 0.05\\
\hline
p min & Runtime  & -& -& -& -& 63.84& 44.41& 36.17\\
= 300 & Speedup  & -& -& -& -& 1.00& 1.44& 1.76\\
 & Load balance  & -& -& -& -& 1.08& 1.10& 1.13\\
 & Search overhead \%  & -& -& -& -& 0.00& 24.81& 81.79\\
 & $R_<$  & -& -& -& -& 0.92& 0.76& 0.55\\
 & $R_=$  & -& -& -& -& 0.08& 0.22& 0.39\\
 & $R_>$  & -& -& -& -& 0.00& 0.02& 0.06\\
 & $R_r$  & -& -& -& -& 0.02& 0.03& 0.05\\
\hline
p min & Runtime  & -& -& -& -& -& 61.58& 41.16\\
= 600 & Speedup  & -& -& -& -& -& 1.00& 1.52\\
 & Load balance  & -& -& -& -& -& 1.09& 1.11\\
 & Search overhead \%  & -& -& -& -& -& 0.00& 29.16\\
 & $R_<$  & -& -& -& -& -& 0.88& 0.70\\
 & $R_=$  & -& -& -& -& -& 0.11& 0.24\\
 & $R_>$  & -& -& -& -& -& 0.01& 0.05\\
 & $R_r$  & -& -& -& -& -& 0.03& 0.05\\
\hline
p min & Runtime  & -& -& -& -& -& -& 58.38\\
= 1200 & Speedup  & -& -& -& -& -& -& 1.00\\
 & Load balance  & -& -& -& -& -& -& 1.09\\
 & Search overhead \%  & -& -& -& -& -& -& 0.00\\
 & $R_<$  & -& -& -& -& -& -& 0.83\\
 & $R_=$  & -& -& -& -& -& -& 0.14\\
 & $R_>$  & -& -& -& -& -& -& 0.03\\
 & $R_r$  & -& -& -& -& -& -& 0.04\\
\hline

\end{tabular}
\caption{24-puzzle scaling summary on the \tsubametwo cluster.
}
\label{tab:tsubame2-24-puzzle-scaling-by-mincores}
\end{scriptsize}
\end{center}
\end{table}
\renewcommand{\arraystretch}{1}

\subsection{Scaling Behavior on Planning on a Commodity Cluster}
\label{sec:hda-scaling-behavior-commodity-cluster}

While the previous set of large-scale experiments were performed on a
campus high-performance computing cluster, we also evaluated the
scalability of HDA* on the \funcluster  cluster\citemachinetable.
Table \ref{hdastar-speed-up-fun-cluster} shows the relative efficiency of 16, 32, and 64 cores compared to a baseline of 8 cores (1 processing node). 
The results are organized according to $p_{min} \in \{8,16,32\}$, the minimum number of tested cores that solved the instance. For each $p_{min}$, the average relative efficiency and relative speedups are shown.
Several trends can be seen. First, when $p_{min}=8$, and the number of cores used is increased to 16, the relative efficiency (0.55) and speedup (1.10) are very poor.
On the other hand, after this initial threshold (the jump from 1 processing node to 2 processing nodes) is crossed, the relative efficiency and relative speedups are near-linear, with low search overheads, similar to our results above on the \tsubameone and \tsubametwo clusters when the number of cores used is within a factor of 8 of $p_{min}$. Inter-node communication plays a role in this behavior. Moving from one processing node to 2 nodes introduces inter-node communication, which is relatively slow in a commodity cluster. From two to more nodes the relative performance grows more steadily, since inter-node communication is present in all multi-node configurations.

In addition, as shown below in Section \ref{sec:hda*-vs-tds}, HDA* significantly outperforms TDS, the previous state of the art algorithm, when the two are compared on this commodity cluster.

\begin{table}[htbp]
\begin{center}
\begin{scriptsize}
\begin{tabular}{|l||r|r|r|r|}
 \hline
        Instance &       HDA* 8 cores & HDA* 16 cores  & HDA* 32 cores  & HDA* 64 cores \\
 \hline
 \hline
 \hline
        Depot10 &   17.98 (1.0) &  16.97 (1.06) &   8.09 (2.22) &   5.60 (3.21) \\

\hline
     Driverlog8 &   17.65 (1.0) &  18.22 (0.97) &   7.29 (2.42) &   4.41 (4.00) \\
\hline
      Freecell5 &   20.82 (1.0) &  16.27 (1.28) &   8.41 (2.48) &   5.99 (3.48) \\
\hline
     Satellite6 &   20.00 (1.0) &  24.64 (0.81) &  19.86 (1.01) &   7.34 (2.72)\\
\hline
      ZenoTrav9 &   27.97 (1.0) &  29.87 (0.94) &  11.65 (2.40) &   7.26 (3.85)\\
\hline
     ZenoTrav11 &   79.43 (1.0) &  81.43 (0.98) &  32.36 (2.45) &  19.33 (4.11)\\
\hline
    PipesNoTk14 &   39.86 (1.0) &  34.26 (1.16) &  12.03 (3.31) &   7.86 (5.07)\\
\hline
       Pegsol27 &   26.97 (1.0) &  22.06 (1.22) &  11.08 (2.43) &   7.43 (3.63)\\
\hline
      Airport17 &   48.84 (1.0) &  33.96 (1.44) &  23.78 (2.05) &  15.99 (3.05)\\
\hline
       Gripper8 &   56.91 (1.0) &  53.95 (1.05) &  22.51 (2.53) &  14.97 (3.80)\\
\hline
       Mystery6 &   48.58 (1.0) &  39.17 (1.24) &  20.13 (2.41) &  11.91 (4.08)\\
\hline
         Truck5 &   69.40 (1.0) &  73.72 (0.94) &  68.71 (1.01) &  22.46 (3.09)\\
\hline
      Sokoban19 &   24.73 (1.0) &  21.10 (1.17) &  11.30 (2.19) &   9.72 (2.54)\\
\hline
      Sokoban22 &   63.65 (1.0) &  49.18 (1.29) &  25.20 (2.53) &  19.57 (3.25)\\
\hline
     Blocks10-2 &   50.83 (1.0) &  47.35 (1.07) &  20.69 (2.46) &  12.64 (4.02)\\
\hline
Logistics00-7-1 &  231.01 (1.0) & 230.97 (1.00) &  91.09 (2.54) &  53.37 (4.33)\\
\hline
{\bf Avg. rel. speedup} & 1.0   &  1.10    & 2.27 & 3.48\\
{\bf Avg. rel. efficiency} & 1.0   &  0.55    & 0.57 & 0.87\\
{\bf Avg. search overhead}  & 0    & 0.20\% & 3.71\% & 4.05\%\\

\hline
        Depot13 &     - (-) & 254.41 (1.0) & 115.85 (2.20) &  75.09 (3.39) \\

\hline
      Freecell7 &     - (-) & 265.00 (1.0) & 128.74 (2.06) &  75.27 (3.52)\\

\hline
        Rover12 &     - (-) & 126.11 (1.0) &  57.66 (2.19) &  40.27 (3.13)\\

\hline
    PipesNoTk24 &     - (-) & 145.49 (1.0) &  63.31 (2.30) &  38.67 (3.76)\\

\hline
       Pegsol28 &     - (-) &  93.55 (1.0) &  45.40 (2.06) &  29.45 (3.18)\\
\hline
       Gripper9 &     - (-) & 273.47 (1.0) & 118.38 (2.31) &  76.39 (3.58)\\
\hline
         Truck6 &     - (-) & 675.71 (1.0) & 337.01 (2.01) & 168.39 (4.01)\\
\hline
         Truck8 &     - (-) & 489.57 (1.0) & 284.01 (1.72) & 116.22 (4.21)\\
\hline
Logistics00-9-1 &     - (-) & 361.88 (1.0) & 154.09 (2.35) &  86.91 (4.16)\\
\hline
   Miconic12-2 &     - (-) & 564.09 (1.0) & 259.01 (2.18) & 159.48 (3.54)\\
\hline
   Miconic12-4 &     - (-) & 600.22 (1.0) & 266.08 (2.26) & 169.49 (3.54)\\
\hline
       Mprime30 &     - (-) & 247.19 (1.0) & 100.27 (2.47) &  68.28 (3.62)\\
\hline
{\bf Avg. rel. speedup} &  -    &  1.0    & 2.18 & 3.63\\
{\bf Avg. rel. efficiency} &  -    &  1.0    & 1.09 & 0.91\\
{\bf Avg. search overhead}  & -    & 0  & 0.61\% & 0.91\%\\
\hline
         Rover6 &     - ( -) &    - ( -) & 268.59 ( 1.0) & 162.82 (1.65)\\

\hline
\multicolumn{5}{|c|}
{Solved only on 64 cores: Freecell6 (302.50 seconds),
Satellite7 (633.26)} \\
\multicolumn{5}{|c|}
{Sokoban26 (205.80),
Blocks11-1 (194.57) and
Logistics00-8-1 (491.02).} \\
\hline
\end{tabular}
\caption{Planning results on the \funcluster cluster: runtime, relative speedup (between brackets), average efficiencies, and average search overhead, by $p_{min}$.
}
\label{hdastar-speed-up-fun-cluster}
\end{scriptsize}
\end{center}
\end{table}

\section{Tuning HDA* Performance}
\label{tuning-hda}

The previous section investigated the scaling behavior of HDA* on
various parallel environments as the amount of available resources
varied.  In this section, we consider how the behavior of HDA* can be
tuned by adjusting two parameters: the number of cores to utilize per
processor node, and the number of states to pack in each message
between processes.

\subsection{Adjusting the Number of Cores to Utilize Per Processing Node}
Next, we investigate the effect of scaling the number of processing nodes (machines) for the 24-puzzle.
We ran the solver on a set of 100 nodes, which have 1200 cores in
total, and varied the number of cores used per node between 1-12, so that 100-1200 cores were used.
The runtimes 
are shown in 
Table~\ref{tab:24-puzzle-100-nodes}.
IDA* solves all 50 instances \cite{KorfF02} whereas with our HDA* using 12 cores 
only 10 instances can be solved with 54GB of memory. 

\renewcommand{\arraystretch}{.88}
\begin{table}[htb]
\begin{center}
\begin{scriptsize}
 \begin{tabular}{|c|c|c|c|c|c|c|c|c|c|c|c|}
\hline
    &  100 cores     &  300 cores  &  600 cores  &  1200 cores\\
\hline
{\bf Average time}& {\bf 209.43} & {\bf 68.61}& {\bf 35.00}& {\bf 35.93}\\
\hline
{\bf Nr solved} & {\bf 45} & {\bf 44} & {\bf 41} & {\bf 40} \\
\hline
\hline
p1 &   6.35 &    4.72 &   15.94 &   35.25\\
\hline
p2 & 356.53 &  116.86 &   75.21 &   45.99\\
\hline
p3 &  31.61 &   12.37 &   15.04 &   30.12\\
\hline
p4 &  33.68 &   14.12 &   14.04 &   28.38\\
\hline
p5 &  11.78 &    7.43 &   13.23 &   30.07\\
\hline
p6 &  75.98 &   28.15 &   24.06 &   36.46\\
\hline
p7 & 211.34 &   75.58 &   48.84 &   42.21\\
\hline
p8 & 190.51 &   74.45 &   51.32 &   34.68\\
\hline
p9 &1038.60 &  365.60 & -         & -        \\
\hline
p12 & 618.90 &  200.37 &  119.41 & -        \\
\hline
p13 &  12.62 &    6.46 &   14.22 &   32.34\\
\hline
p15 & 446.57 &  167.86 &   99.71 &   57.72\\
\hline
p16 &  10.77 &    5.71 &   11.20 &   26.67\\
\hline
p18 &1032.34 &  367.38 & -         & -        \\
\hline
p19 & 162.77 &   58.76 &   40.16 &   37.02\\
\hline
p20 & 188.30 &   69.20 &   47.09 &   33.40\\
\hline
p21 & 925.68 &  333.59 & -         & -        \\
\hline
p22 &   4.12 &    4.60 &   12.75 &   24.58\\
\hline
p23 & 410.70 &  156.23 &   96.00 &   61.63\\
\hline
p24 & 277.98 &   92.09 &   58.31 &   52.59\\
\hline
p25 &   1.37 &    3.44 &   10.19 &   19.30\\
\hline
p26 &  34.32 &   13.88 &   18.86 &   47.11\\
\hline
p27 &  68.71 &   24.33 &   23.68 &   35.20\\
\hline
p28 &   2.55 &    5.33 &   13.33 &   26.13\\
\hline
p29 &   8.23 &    5.61 &   11.80 &   24.51\\
\hline
p30 &   4.54 &    4.38 &   11.91 &   25.22\\
\hline
p31 &  52.41 &   20.46 &   19.12 &   34.28\\
\hline
p32 &   2.36 &    4.78 &   10.85 &   29.31\\
\hline
p33 &1104.11 & -         & -         & -        \\
\hline
p34 & 412.80 &  154.47 &   91.31 &   57.49\\
\hline
p35 & 183.70 &   65.78 &   50.36 &   33.55\\
\hline
p36 &   4.70 &    4.70 &   12.24 &   31.58\\
\hline
p37 &  10.17 &    6.36 &   14.28 &   30.20\\
\hline
p38 &   1.57 &    5.36 &   13.48 &   28.11\\
\hline
p39 & 280.13 &   94.78 &   57.25 &   44.66\\
\hline
p40 &   1.01 &    3.44 &   10.21 &   22.02\\
\hline
p41 &  41.20 &   17.07 &   16.88 &   37.99\\
\hline
p42 & 282.17 &   98.58 &   56.30 &   42.23\\
\hline
p43 &  52.75 &   20.57 &   20.61 &   37.44\\
\hline
p44 &   1.48 &    4.06 &   11.88 &   24.89\\
\hline
p45 & 132.27 &   43.76 &   32.06 &   34.47\\
\hline
p46 & 163.49 &   59.35 &   42.30 &   37.01\\
\hline
p47 &  32.70 &   13.52 &   16.29 &   36.97\\
\hline
p48 & 412.32 &  152.31 &   90.36 &   56.71\\
\hline
p49 &  86.03 &   30.95 &   22.88 &   31.61\\
\hline
\end{tabular}
\caption{Runtimes in seconds for the 24-puzzle on 100 nodes of the \tsubametwo cluster.
Average runtime only includes instances solved by all configurations.
Instances which were not solved by any configuration are excluded. 
}
\label{tab:24-puzzle-100-nodes}
\end{scriptsize}
\end{center}
\end{table}
\renewcommand{\arraystretch}{1}

As shown in
Table~\ref{tab:24-puzzle-100-nodes},
using 1200 cores (4.5GB/core), 40 out of the 50 problems were
solved. With 600 cores (9GB/core), 41 problems were solved.  With 300
cores (18GB/core), 44 problems were solved, and with 100 cores
(54GB/core), 45 problems were solved. That is, {\em reducing the number of
processes down to one process per node increases the number of solved
instances}. 

A closer look at the trade-offs involved explains this behavior.
First there is the reduction in execution time when performing a given amount of work with more cores.
Indeed, Table~\ref{tab:24-puzzle-100-nodes} indicates that, \emph{if} an instance is solved
by a larger number of cores, the time tends to decrease as more cores were used.
The notable exception is the configuration with 1200 cores where
the large search overhead actually degrades the time performance.
Other exceptions are instances that are easy for 100 cores, such as
p22, p25, p28, p38, p40 and p44, which were solved in a few seconds and left little room for further improvement in runtime, resulting in consistently increasing runtimes as the number of cores increased.
On the other hand, using more cores per processing node can lead to
solving fewer instances.
Suppose that $k$ unique states need to be stored in
open/closed in order to solve a problem. A sequential search with enough 
memory capacity to hold $k$ states can solve this problem. 
If HDA* allocated work perfectly equally among the processors, and there was no search overhead,
then the partitioning of memory among $n$ cores would have no negative impact.
In practice, load balancing is imperfect, and search overhead is non-zero, 
so increasing the number of cores (for a fixed amount of aggregate RAM)
increases the chance of failure on a hard problem.

When analyzing the relative impact of search overhead vs. RAM partitioning,
we found that
the former plays a significantly greater role in the reduction of the 
number of solved instances.
Table~\ref{tab:tsubame2-24-puzzle-scaling-by-mincores} shows that
the search overhead eventually increases
significantly as the number of cores increases, meaning that more nodes must
be stored in the open/closed lists.  
The rate of increase accelerates as the number of cores increases further beyond the minimal number of cores needed to solve the problem.
On the other hand, our use of static RAM partitioning (as opposed to a more flexible, dynamic partitioning) is not a significant bottleneck.
We measured the size of the open and closed lists
for all processors, and observed that 
when the first processor runs out of memory,
most other processors have almost exhausted their memory as well.
This means that there is little opportunity for improvements to be made by using a  more flexible, dynamic memory partitioning method.

In Section \ref{sec:scale-nodes}, we observed slowdowns in
domain-independent planning as the ratio of utilized cores per node increased,
and ascribed the slowdown to local memory bus contention.
In the 24-puzzle, the node expansion rate 
decreases by 35\% as the number of cores per processing node increased from 1 to 12.
We attribute this increase to both local memory bus contention, and factors
discussed in Section~\ref{node expansion rate}.
\subsection{The Effect of the Number of States Packed into Each Message}
\label{sec:packing-states}
In HDA*, each state generation necessitates sending the state from
processor where a state is generated to the processor which ``owns''
the state. Sending a message from processor $P$ to $Q$ each time a
state owned by $Q$ is generated at $P$ may result in excessive
communication overhead, as well as overhead for creating/manipulating
MPI message structures.  

In order to amortize these overheads, Romein
et al. \cite{Romein:99}, in their work on TDS, proposed packing
multiple states with the same destination.

On the other hand, packing
too many states into a message from processor $P$ to $Q$ might result
in degraded performance for two reasons. First, the destination $Q$
might be starved for work and be idle.  Second, too much packing can
result in search overhead, as follows. Consider a state $S$ on an
optimal path, which is ``delayed'' from being sent to its owner
because the processor which generated $S$ is waiting to pack more
states into the message containing $S$. In the meantime, the owner of
$S$ is expanding states which are worse than $S$ and possibly sending
those successors to fill up the open list of their owner processors, and so on.

We compared packing 10, 100, and 1000 states per message on the
\funcluster cluster using 64 cores, using the same benchmark instances
in Table \ref{hdastar-speed-up-fun-cluster}, except that the Airport17
instance was excluded because it could not be solved using 10 states
per message.
Using 10 states per message, the average runtime was 31.6\% slower than when using 100 states per message, and 
1.0\% fewer nodes were expanded.
Using 1000 states per message, the average runtime was 37.4\% slower than when using 100 states per message, and 11.6\% more nodes were expanded. 
Thus, while packing fewer states per message reduces search overhead, packing more states per message reduces communication overhead, and in this case, packing 100 states per message performs well by striking a balance between these two factors.

\section{Hash-Based Work Distribution vs. Random Work Distribution on a HPC Cluster}
\label{sec:hda-random-distribution-results}

Kumar et al. \cite{Kumar:1988} and Karp and Zhang \cite{KarpZ88}
proposed a simple, random work distribution strategy for best-first
search where generated nodes are sent to a random processor.  
While
this is similar to HDA* in that a randomization mechanism is used to
distribute work, the difference is that duplicate states are not
necessarily sent to the same processor, since a state has no ``owner''.
Although duplicates are pruned {\em locally} at each processor, there is no global duplicate detection, so 
in the worst case, a state
can be in the local open/closed lists of every single processor.

We evaluated this ``Random'' work distribution strategy 
on our Fast-Downward based domain-independent planner
on the \tsubameone cluster.
First, we attempted to compare HDA* and Random using 16 cores, 2GB per core.
However, the Random strategy failed to solve any of the test problems in this configuration -- all of the runs failed due to memory exhaustion.
This indicated that, due to the lack of (global) duplicate detection, the random work distribution strategy requires much more RAM to solve our benchmarks.
Hence we performed a comparison using 16 cores, 32GB per core (as with the experiments in Section \ref{sec:scale-nodes}, this configuration left 15 of the 16 cores on each processing node idle, in order to maximize memory available per core).
The results are shown in Table \ref{tab:random-distribution}. The execution time and number of nodes expanded by HDA* is more than an order of magnitude less than those of the random work distribution strategy.
In fact, the random work distribution strategy is {\em slower} than the sequential A* algorithm because of large search overhead.
The load balance is similar for both HDA* and random work distribution, indicating that hash-based work distribution is successfully distributing the work as evenly as a pure, randomized strategy.
These results clearly demonstrate the benefit of using hash-based work
distribution in order to perform global duplicate detection as well as
load balancing.

\begin{table}[htb]
\begin{center}
\begin{footnotesize}
 \begin{tabular}{|l|r|r|r|r|r|r|r|r|}
\hline
&          \multicolumn{2}{|c|}{Execution time} & \multicolumn{2}{|c|}{nodes expanded $\times10^6$} & \multicolumn{2}{|c|}{load balance} \\
\hline
           & Random            & HDA*                  & Random            & HDA*                 & Random            & HDA*      \\
\hline
Driverlog13 & n/a               & 1056.24              & n/a               & 618      & n/a               & 1.05 \\
\hline
Freecell6   & n/a               & 990.19               & n/a               & 341            & n/a               & 1.04 \\
\hline
Freecell7   & 3170.81           & 234.37               & 1,202        & 97             & 1.01              & 1.02 \\
\hline
Satellite7  & n/a               & 1493.33              & n/a               & 219           & n/a               & 1.005 \\
\hline
ZenoTrav11  & 1022.57           & 54.93                & 221         & 14             & 1.0002            & 1.13 \\ 
\hline
ZenoTrav12  & n/a               & 4393.80              & n/a               & 1,152           & n/a               & 1.03 \\
\hline
PipesNoTk24 & 2094.81           & 130.13               & 1,065        & 72             & 1.0001            & 1.005 \\
\hline
Pegsol28    & 1040.28           & 74.04                & 1,364        & 112            & 1.01              & 1.005 \\
\hline
Pegsol29    & n/a               & 364.39               & n/a               & 481            & n/a               & 1.01 \\
\hline
Pegsol30    & n/a               & 978.11               & n/a               & 1,379           & n/a               & 1.01 \\
\hline
Sokoban24   & 2748.71           & 171.28               & 3,254        & 244            & 1.01              & 1.01 \\
\hline
Sokoban26   & n/a               & 488.15               & n/a               & 659            & n/a               & 1.01 \\
\hline
Sokoban27   & n/a               & 942.57               & n/a               & 1,162           & n/a               & 1.03 \\
\hline

\end{tabular}
\caption{HDA* vs random work distribution (``Random'').
Execution time, nodes expanded, and load balance on 16 cores (on 16 processing nodes), using 32GB per core.
Experiments were performed on the \tsubameone cluster.
}
\label{tab:random-distribution}
\end{footnotesize}
\end{center}
\end{table}

\section{Comparison of HDA* vs.  TDS on a Commodity Cluster}
\label{sec:hda*-vs-tds}

Transposition-Driven Scheduling (TDS) is a parallelization of IDA*
with a distributed transposition table, where hash-based work
distribution is used to map states to processors for both scheduling
and transposition table checks \cite{Romein:2002}.  TDS has been
applied successfully to sliding tiles puzzles and Rubik's Cube
\cite{Romein:2002}, and has also been adapted for adversarial search
\cite{Kishimoto:02,Romein:2003}.  While TDS has not been applied to
planning, recent work has shown that IDA* with a transposition table (IDA*+TT)
is a successful search strategy for optimal, domain-independent
planning \cite{AkagiKF2010} -- on problems which can be solved by A*, the runtime of IDA*+TT is usually within a factor of 4 of A*, and IDA*+TT can eventually solve problems where A* exhausts memory.  Therefore, we compared HDA* and TDS
\cite{Romein:2002} for planning.

The experiments were performed on 
the \funcluster cluster{\citemachinetable} using 64 cores, with a 20 minute time limit per instance.
Both HDA* and TDS were run on 35 planning instances (the same instances as for the experiments in Section \ref{sec:hda-scaling-behavior-commodity-cluster}, Table \ref{hdastar-speed-up-fun-cluster}).
 
Our implementation of TDS uses a transposition table
implementation based on \cite{AkagiKF2010}, where the table entry
replacement policy is a batch replacement policy which sorts entries according to access frequency and periodically frees 30\% of the entries (preferring to keep most frequently accessed entries).\footnote{Although
  replacement based on subtree size performed best in
  sequential search \cite{AkagiKF2010}, we did not implement this policy because subtree
  size computation would require extensive message passing in parallel search. 
} 
We incorporated techniques to overcome higher latency in a lower-bandwidth network
described in \cite{Romein:2002}, such as their modification to the termination detection
algorithm. 

\begin{figure}
\begin{center}
\includegraphics[width=.6\textwidth]{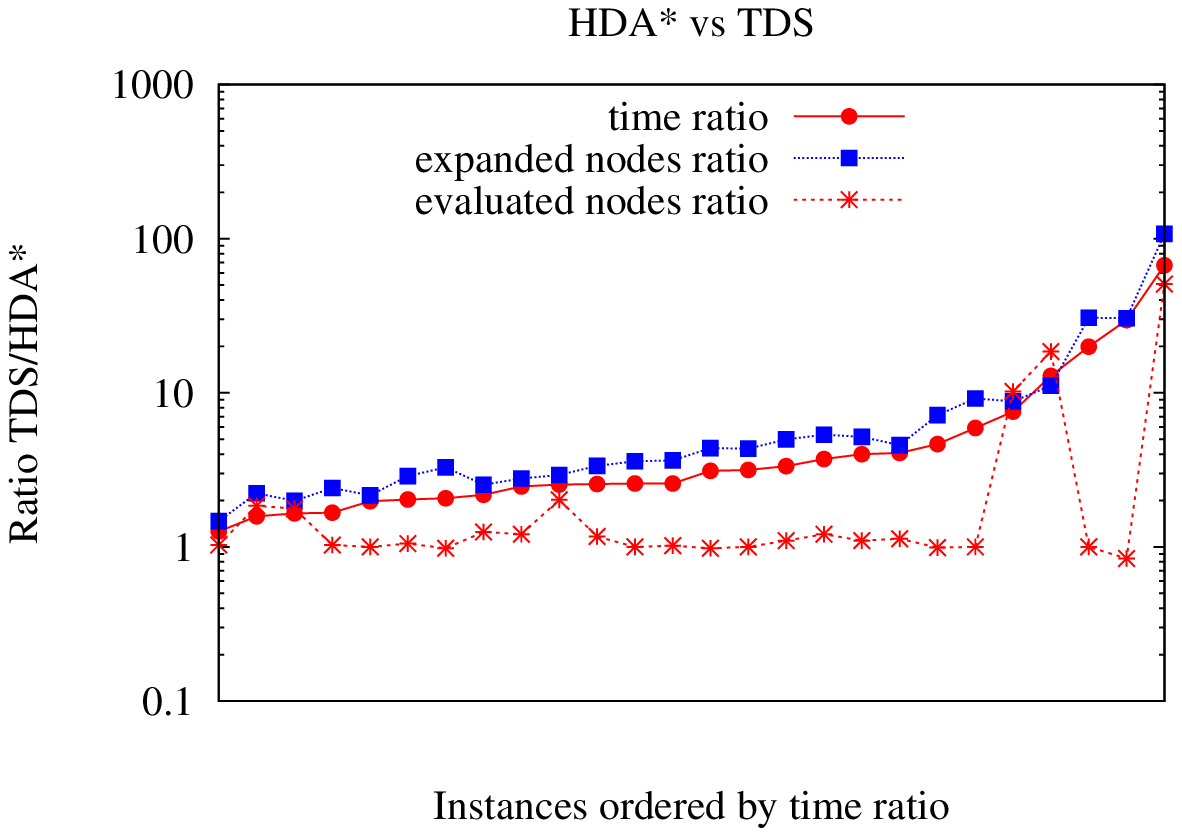}
\end{center}
\caption{Ratios TDS/HDA* on 64 cores, on the \funcluster cluster\citemachinetable.}
\label{fig-ratios}
\end{figure}

Out of 35 instances, HDA* failed to solve one
case (Blocks12-1) and TDS failed to solve 8 cases (Logistics00-8-1, Sokoban22, Sokoban26, Truck6, Gripper9, Freecell6, Rover6, Satellite7).
Figure~\ref{fig-ratios} directly compares HDA* and TDS
on 26 instances solved by both algorithms, 
for which it is possible to compute the ratios of performance measures such as
time, expanded states and evaluated states.\footnote{Evaluated states are states which were not found in the Open/Closed set (in the case of  HDA*), or not found in the transposition table (in the case of TDS).}
HDA* is consistently faster, with a maximum speedup of about 65.
There are several differences between HDA* and TDS that 
could cause this significant performance difference.
Below, we first enumerate these  differences, and then 
consider how they apply to our results. %

First, TDS is an iterative deepening strategy, so TDS will incur
state reexpansion overhead, since many states will be reexpanded as
the iteration $f$-bound increases.

Second, HDA* opens states in a (processor-local) best-first order, while 
TDS opens states in a last-in-first-out order in its local stack,
subject to an iteration bound for the $f$-value.
This difference in state expansion policy results in  
different search overheads 
for the two algorithms.
Third, each processor in TDS  needs memory not only 
for a transposition table and the merge-and-shrink abstraction (for planning),
but also for a work stack.
In the tested configuration, %
each processor is allocated 2GB of memory. 
1.2GB is allocated to the transposition table and the abstraction.
The remaining memory is reserved for the work stack.\footnote{While Romein et al.'s stack does not exceed 1MB in their
applications \cite{Romein:2002}, our stack often used hundreds of megabytes of memory. 
Due to the much larger branching factors of our planning domains, compared to their domains (15-puzzle, Rubik's cube), TDS sends away more states when generating successors, causing a larger work queue.
A possible future improvement is to combine Dutt and Mahapatra's technique 
in SEQ\_A* \cite{MahapatraD97} with TDS and restrict initiating parallelism.}
Therefore, there may be 
cases where there is sufficient aggregate memory in HDA* to solve an
instance, but TDS (on the same system) exhausts the space allocated
for its distributed transposition table. In such a situation, TDS
applies a replacement policy to replace some of its transposition
table entries with newly expanded states.  While this seeks to maintain the most useful working set of states, it is possible that
valuable entries (states) in the table are replaced, resulting in
wasted work later when these states are revisited. 

Fourth, TDS incurs synchronization overhead while all processors wait for the current
iteration to end. 

Standard IDA* often generates states faster than 
A*,  because  successor generation can be implemented as an in-place incremental update of a data structure representing the current state.
Thus, sequential IDA* can outperform A* even if IDA* searches less efficiently than A*.
When a transposition table is added to
IDA*, state generation incurs a significant overhead because the hash
value of each generated state must be computed, and if there is a
transposition table miss, some hashed representation of the state must
be stored in the table. 
Furthermore, in TDS, {\em every} distributed transposition table access
requires a hashed representation of the state to be generated and sent
to the owner process. Thus, parallel IDA* no longer
has an inherently faster state generation rate than parallel A*.

We now consider the contribution of each of the above differences.
In Figure~\ref{fig-ratios}, the ratio of expanded states
closely follows the runtime ratio.\footnote{
The ratio of generated states, not shown to reduce clutter, is almost identical to
the ratio of expanded states.}
On the other hand, the ratio of evaluated states is close to 1 in most cases.
This %
suggests that both algorithms explore a similar number of
\emph{unique} states.
Furthermore, the similarity in the number of evaluated
states shows that the transposition table used by TDS is large enough
to fit most (unique) states generated during search, so the size of
the transposition table did not limit the performance of TDS. %

The difference in performance is mostly due to the re-expansion overhead incurred by TDS during iterative deepening.
This conclusion is supported by the close correspondence between the runtime ratio and the state expansion ratio. 
Also, it is consistent with the behavior of sequential A* and IDA*.
For example, on the Airport17 instance, 
both A* and IDA* evaluate 10,397,245 states.
While A* expands only 7,126,967 states,
IDA* expands 239,705,187 states, indicating that most of the
states are reexpanded.\footnote{
Extending TDS 
to backpropagate search results
might reduce the very high reexpansion overhead in planning, 
which did not occur in Romein et al.'s applications.}

\begin{figure}
\begin{center}
\includegraphics[width=.7\textwidth]{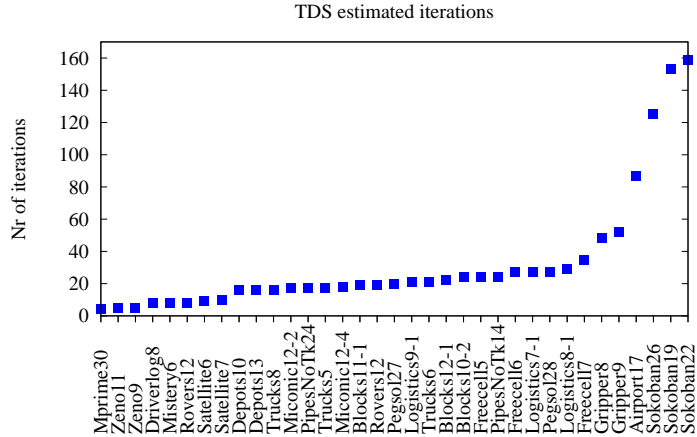}
\end{center}
\caption{Estimated number of TDS iterations.
}
\label{fig-tds-iterations}
\end{figure}

Clearly, the number of re-expansions depends on the number of iterations performed by TDS.
The exact number of iterations is unavailable when TDS runs out of time.
An upper bound can be computed as $u = c^* - h(s_0) + 1$, where $c^*$ is the optimal cost (available due to HDA*),
and $h(s_0)$ is the heuristic evaluation of the initial state. 
We observed that this was exactly the number of iterations in the cases where TDS succeeded within the time limit.
More generally, in all cases that we observed, the cost threshold increase
from one iteration to the next was 1.
Thus, we hypothesize that the upper bound $u$ is in fact quite an accurate estimate
of the actual number of iterations.
On the 35 instances considered in this experiment, the (estimated) number of iterations varies from 4 
to 159, with an average value of 32.48. Data for all instances are available in Figure~\ref{fig-tds-iterations}.

The synchronization overhead in TDS between iterations, as all processors wait for the iteration to end, does not appear to be significant in most cases, which is consistent with Romein et al.'s results  (otherwise, the runtime ratio would be consistently higher than the state expansion ratio).

\begin{figure}
\begin{center}
\includegraphics[width=.6\textwidth]{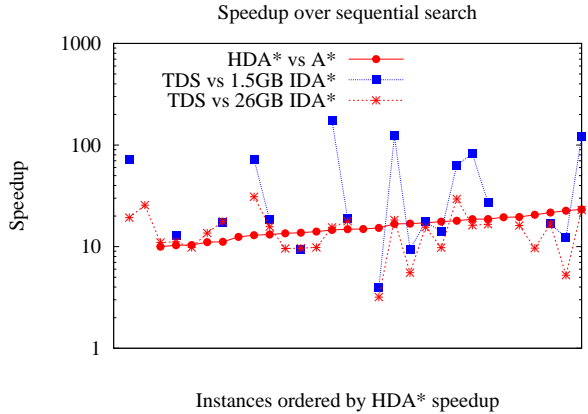}
\end{center}
\caption{Speedup over sequential search. A missing data point indicates that the 
sequential algorithm or both the sequential and the parallel algorithm failed to find a solution.
Experiments were performed on the \funcluster cluster.
}
\label{fig-speedups}
\end{figure}

Next we consider the speedups obtained by HDA* and TDS over their sequential counterparts,
A* and IDA* with transposition table (IDA*+TT).
The speedup of TDS vs IDA*+TT depends greatly on the amount of RAM memory available for the
transposition table.
The larger the (aggregate) transposition table, the faster TDS and IDA* are.
While it is possible to achieve drastically large (sometimes super linear) speedup over IDA*~\cite{Romein:99,Romein:2002},
the speedup diminishes as IDA* has access to larger and larger transposition tables.
Figure~\ref{fig-speedups} shows that, depending on the amount of RAM
available to IDA*+TT, the speedup of TDS vs IDA*+TT can be either smaller or larger
than speedup of HDA* vs A*.
In this experiment, serial IDA*+TT was tested using 1.5GB of RAM and 26GB of RAM.
TDS uses 1.2GB $\times$ 64 cores, or 76.8GB of aggregate RAM. 
The time limit of serial IDA*+TT was set to 1280 minutes (i.e., about 21.3 hours) 
per instance. 

\subsection{A Simple, Hybrid Strategy Combining HDA* and TDS}
\label{sec:hybrid}

The results above indicate that, for planning, HDA* significantly
outperforms TDS on instances that can be solved within the memory available to HDA*.
However, while HDA* will terminate
and fail when memory is exhausted at any processor,
TDS will not terminate if
the local transposition table at any processor becomes full. 
Mechanisms which allow A*-based search to continue after memory is
exhausted have been proposed.
For example, PRA* \cite{Evett:1995},
has a state retraction mechanism which frees memory by
retracting some states at the search frontier (however, as explained in Section \ref{sec:background}), the PRA* retraction policy results in synchronization overhead).  
Determining an effective state retraction policy is a non-trivial extension to HDA* and an interesting direction for future work.

Consider instances which are solvable by TDS (within a reasonable amount of time) but not by HDA*.
Such problems are not very common.
Given that HDA* and TDS explore a very similar set of unique states (as shown above),
if a problem cannot be solved by HDA*, it indicates that the instance is quite
difficult, and it is likely that TDS cannot solve the problem either
within a reasonable time limit.

The instances that HDA* failed to solve, as well as the amount of time HDA* executed before exhausting RAM) are listed below for 16, 32, and 64 cores (all with 2GB per core):

\begin{itemize}
\item Failed with 16 cores: Freecell6 (269 sec), Rover6 (453 sec), Satellite7 (573 sec), Sokoban26 (176 sec), Blocks11-1 (165 sec), Blocks12-1 (148 sec), Logistics00-8-1 (416 sec);
\item Failed with 32 cores: Freecell6 (303 sec), Satellite7 (722 sec), Sokoban26 (190 sec), Blocks11-1 (190 sec), Blocks12-1 (156 sec), Logistics00-8-1 (399 sec);
\item Failed with 64 cores: Blocks12-1 (189 sec).
\end{itemize}

Next, we attempted to solve these instances using TDS with 16, 32, and
64 cores (all with 2GB/core), with an extended time limit of 1.5 hours 
for 32 and 64 cores and 3 hours for 16 cores 
per instance. The runtimes on the instances which were
solved by TDS, but not by HDA* are as follows (TDS failed to find a solution within
the time limit on the other instances):

\begin{itemize}
\item 16 cores:  Blocks11-1 (2928 sec), Blocks12-1 (4139 sec);
\item 32 cores:  Freecell6 (4350 sec), Blocks11-1 (974 sec), Blocks12-1 (1874 sec); 
\item 64 cores:  Blocks12-1 (838 sec). 
\end{itemize}

Thus, even with a 1.5 or 3 hours per instance, many
instances that cannot be solved by HDA* cannot be solved by TDS.  In
addition, on the instances where HDA* fails and TDS succeeds, HDA*
fails relatively quickly compared to the time required by TDS to solve
the problem. For example, with 32 cores HDA* exhausts memory in 156
seconds in solving Blocks12-1, and although TDS can solve this problem with 32 cores, it
required 1874 seconds.

This suggests a very simple, hybrid approach which 
combines the speed of HDA* and the ability of
TDS to eventually solve difficult problems without running out of memory.
First, a slightly modified version of HDA* is applied, which keeps track of
the lowest $f$-cost in the OPEN list is recorded at every processor.
If HDA* finds a solution, then it is returned. However, if HDA*
exhausts memory at any processor, then it terminates with failure, and
also returns the minimum frontier value among all of the processors,
$f_{min}$.  Then, we apply 
TDS, except that instead of setting the initial iteration bound to 
the heuristic value at the initial state, 
we
start with $f_{min}$ as the lower bound.

According to the data above,
if the HDA* phase succeeds, this hybrid procedure will succeed significantly faster than TDS would,
and if it fails, it will fail relatively quickly. Upon failure, the
hybrid starts TDS with a iteration bound $f_{min}$ (skipping some wasteful iterations).

On instances solvable by HDA*, the hybrid runtime 
will be the same as for HDA*.  On instances where HDA* fails
but TDS would eventually succeed, the hybrid runtime would be
similar to the runtime of TDS. 
The time spent in the failed run of
HDA* should only be a small fraction of the time required to eventually
solve the instance; furthermore, using the $f_{min}$ initial bound for
TDS is expected to offset some of the time spent in the failed HDA* run by eliminating 
some of wasted expansions in TDS.

Table~\ref{tds-vs-hybrid} shows the results of applying this hybrid to the instances listed above which caused HDA* to fail.
In most cases, the runtimes for the hybrid are
comparable to the runtimes for TDS.%
This shows that the hybrid successfully combines the strengths of both HDA* and TDS while avoiding their disadvantages, and does so while avoiding excessive hybridization overhead.

\begin{table}[htb]
\begin{center}
\begin{tabular}{|l||r|r|r||r|}
\hline
         & \multicolumn{3}{|c||}{Hybrid} &  \\
Instance & HDA* & TDS & Total & TDS \\
\hline
\hline
\multicolumn{5}{|l|}{\bf 16 cores} \\
Blocks11-1 & 165 & 2517 & 2682 & 2928 \\
\hline
Blocks12-1 & 148 & 4192 & 4340 & 4139 \\
\hline
\hline
\multicolumn{5}{|l|}{\bf 32 cores} \\
Freecell6  & 303 & 3781 & 4084 & 4350 \\
\hline
Blocks11-1 & 190 & 865  & 1055 & 974 \\
\hline
Blocks12-1 & 156 & 1876 & 2032 & 1874 \\
\hline
\hline
\multicolumn{5}{|l|}{\bf 64 cores} \\
Blocks12-1 & 189 & 566 & 755 & 838 \\
\hline
\end{tabular}
\caption{Time in seconds for TDS and Hybrid HDA* + TDS (these are instances which could not be solved by HDA* by itself).
Experiments were performed on the \funcluster cluster.
}
\label{tds-vs-hybrid}
\end{center}
\end{table}

\section{Related Work}
\label{sec:related-work}

We have discussed the related work that parallelizes search by partitioning the search space in Section \ref{sec:background}.
In this section, we review related work on parallel search in model checking.
In the second half, we also review other approaches to parallelizing search algorithms.

While this paper focuses on standard AI search domains including domain independent planning and the sliding tile puzzle,
distributed search, including hash-based work distribution, has also been studied extensively by the parallel model checking community.
Parallel Mur$\varphi$~\cite{SternD97,SternD01} addresses verification tasks
that involve exhaustively enumerating all reachable states in a state space.
Similarly to HDA* and other work described in Section \ref{sec:background}~\cite{Evett:1995,MahapatraD97}, 
Parallel Mur$\varphi$ implements a hash-based work distribution schema where each state is 
assigned to a unique owner processor. 
Kumar and Mercer~\cite{Kumar05loadbalancing} present a load balancing technique
as an alternative to the hash-based work distribution implemented in Mur$\varphi$.
The Eddy Murphi model checker~\cite{Melatti:2009} specializes processors' tasks,
defining two threads for each processing node. The worker thread performs state processing
(e.g., state expansion), whereas the other thread handles  communication (e.g., sending and receiving states). 

Lerda and Sisto parallelized the SPIN model checker to increase the 
availability of memory resources~\cite{lerda-99}.
Similar to hash-based distribution,
states are assigned to an owner processing node, and get expanded at their owner node.
However, instead of using a hash function to determine the owner processor,
only one state variable is taken into account.
This is done to increase the likelihood that the processor where a state is
generated is identical to the owner processor.
Holzmann and Bo\^{s}na\^{c}ki~\cite{HolzmannB07} introduce an extension
to SPIN to multicore, shared memory machines. 
Garavel et al.~\cite{GaravelMS01} use hash-based work distribution to convert an implicitly defined
model-checking state space into an explicit file representation.
Symbolic parallel model checking has been addressed in~\cite{DBLP:conf/cav/HeymanGGS00}.

Thus, hash-based work distribution and related techniques for distributed search have been widely studied for parallel model checking.
There are several important differences between previous work in model checking and this paper.
First, this paper focuses on parallel A*. In model checking, there is usually no heuristic evaluation function, 
so depth-first search
and breadth-first search is used instead of best-first strategies such as A*.

Second, reachability analysis in model checking
(e.g.,~\cite{SternD97,SternD01,lerda-99,GaravelMS01}),
which involves visiting all reachable states,
does not necessarily require optimality.
Search overhead is not an issue because 
both serial and parallel solvers will expand all reachable states exactly once.
In contrast, we specifically address the problem of finding an optimal path, a significant constraint which introduces the issue of search efficiency because distributed A* (including HDA*) searches many nodes with the $f$-cost greater than or equal to the optimal cost, as detailed in Section~\ref{so-planning}; 
furthermore, node re-expansions in parallel A* can introduce search overhead. 
This is the first paper to analyze search overhead.

Finally, while previous work in model checking has used up to 256 processors \cite{VBBB09}, 
our work presents the largest scale experiments with hash-based work distribution to date, showing that hash-based work distribution can scale efficiently relative to $p_{min}$ even for up to 2400 processors.

Kobayashi et al.~\cite{KobayashiKW2011} have applied HDA* to multiple sequence alignment (MSA).
The non-unit transition costs lead to a higher rate of re-expansions, increasing the search overhead.
To address this, the authors introduced a work distribution strategy that 
exploits the structure that MSA and potentially other problems exhibit,
replacing the Zobrist-based, global hashing scheme.
In contrast, in this work, our focus is to analyze HDA*	at length, hoping to establish
HDA* as a simple and scalable baseline algorithm for parallel optimal search.

One alternative to partitioning the search space among processors is to parallelize the computation done during the
processing of a single search node (c.f., \cite{CampbellHH02, CazenaveJ08}).  
The Operator Distribution Method for parallel Planning (ODMP) \cite{VrakasRV01}
parallelizes the computation at each node.
In ODMP, there is a single, {\em
  controlling thread}, and several {\em planning threads}.  The controlling
thread is responsible for initializing and maintaining the current
search state. At each step of the controlling thread main loop, 
it generates the applicable operators, inserts them in
an {\em operator pool}, and activates the planning threads. Each
planning thread independently takes an operator from this shared
operator pool, computes the grounded actions, generates the
resulting states, evaluates the states with the heuristic function,
and stores the new state and its heuristic value in a {\em global agenda} 
data structure. %
After the operator pool is empty,  %
the controlling thread extracts the best new
state from the global agenda, assigning it to the new, current
state. %

Another approach is to run a set of different search algorithms in parallel.
Each process executes mostly independently, with
periodic communication of information between processors.  This
approach, which is a parallel version of an {\em algorithm portfolio}
\cite{HubermanLH97}, seeks to exploit the long-tailed runtime
distribution behavior encountered in search algorithms
\cite{GomesSCK00} by using different versions of search algorithms to
search different (potentially overlapping) portions of the search space. 
An example of this is
the ManySAT solver \cite{HamadiJS09}, which executes
a different version of a DPLL-based backtracking
SAT solver on each processor and periodically shares lemmas
among the processes. 

A third approach is parallel-window search for IDA* \cite{PowleyK91}, where
each processor searches from the same root node, 
but is assigned a different
bound -- that is, each processor is assigned a different, independent iteration of
IDA*. When a processor finishes an iteration, it is assigned the next
highest bound which has not yet been assigned to a processor.  

Vidal et al.~\cite{VBH:socs2010} propose a multicore version of the 
KBFS algorithm~\cite{Felner03kbfs:k-best-first}.
In this approach, each thread expands one node from the Open list 
at a time. 
As each expansion step requires operations on the Open and Closed list,
synchronization is needed to ensure that only one thread at a time can
perform such operations.
Experiments are reported for satisficing planning, as opposed to our work,
which is focused on optimal planning.

The best parallelization strategy for a search algorithm depends on
properties of the search space, as well as the parallel architecture on
which the search algorithm is executed. The EUREKA system
\cite{CookV98} used machine learning to automatically configure
parallel IDA* for various problems (including nonlinear planning) and machine architectures.

Niewiadomski et al. \cite{Niewiadomski06} propose PFA*-DDD, a parallel version of 
Frontier A* with Delayed Duplicate Detection.
While achieving very good speed-up efficiency (often superlinear), 
PFA*-DDD is limited to returning only the cost of a path from start to target,
not an actual path. 
While divide-and-conquer (DC) can be used to reconstruct a path (as in sequential frontier search), parallel DC poses non-trivial design issues that need to be addressed in future work. See the original paper for a discussion.

\section{Discussion and Conclusion}
\label{sec:discussion}

This paper investigated the use of hash-based work distribution to
parallelize A* for hard graph search problems such as domain-independent
planning.  
We implemented Hash-Distributed A*, 
a simple, scalable parallelization of A*.
The key idea, first used in Parallel Retracting A* \cite{Evett:1995}, is to 
distribute work according to the hash value for generated states.
HDA* is a simple implementation of this idea, which, to our knowledge,
has not been previously evaluated in depth.
Unlike PRA*, which was a synchronous algorithm due to its retraction mechanism, HDA* operates completely asynchronously.
Also, unlike previous work such as PRA* and GOHA \cite{MahapatraD97}, which implemented hash-based work distribution on variants of A*, HDA* is a straightforward implementation of hash-based work distribution for standard A*.
We evaluated HDA* as a replacement for the
sequential A* search engine for a state-of-the-art, optimal
sequential planner, Fast Downward %
\cite{Helmert:07}. 
We also evaluated HDA* on the 24-puzzle domain by implementing a parallel solver 
with a disjoint pattern database heuristic \cite{KorfF02}.%

Our experimental evaluation shows that HDA* scales well in several
parallel hardware configurations, including a single shared memory
machine, high-performance computing clusters using Infiniband
interconnects, and a local cluster with 1Gb(x2) Ethernet network.

While HDA* is naturally suited for distributed memory parallel search on a
distributed memory cluster of machines, we have shown that HDA* also achieves reasonable
speedup on a single, shared memory machine with up to 8 cores, 
yielding speedups of 3.8-6.6 on 8 cores.
Burns et al. have recently proposed PBNF, a shared memory, parallel
best-first search algorithm, and showed that PBNF outperforms HDA* on
planning in shared memory environments \cite{BurnsLRZ10}.
On the other hand, while HDA* is not necessarily the fastest algorithm on
a shared memory environment, HDA* is simpler than PBNF
(which is based on structured duplicate detection techniques).
Furthermore, HDA* is suited for larger, distributed memory clusters, while PBNF is designed for shared memory machines.

Evaluation of HDA* on a large high-performance cluster using up to 2400 cores showed that 
HDA* scales well relative to $p_{min}$, the minimum number of processors (+memory) that can solve the problem. In our benchmarks $p_{min}$ ranged up to 600 cores.
When up to 8 times $p_{min}$ processors are used, HDA* scales up relatively efficiently. 
However, as additional processors are used, the search overhead (wasteful node expansions) increases, resulting in degraded efficiency as the number of processors used greatly exceeds $p_{min}$.
Comparison with a randomized work distribution strategy which performs load balancing but no duplicate detection
\cite{Kumar:1988,KarpZ88} showed the simple hash-based duplicate detection mechanism is essential
to the performance of HDA*.
The scaling behavior of HDA* on a small commodity cluster with 1Gb(x2) Ethernet shows similar trends as on the HPC cluster, %
except that going from 8 cores (1 processor node) to 16 cores (2 processor nodes) is inefficient. Increasing the number of processing nodes beyond 2 scaled smoothly. %
Qualitatively similar scaling results were obtained for the 24-puzzle. However, because of the faster node processing time for the 24-puzzle compared to domain-independent planning, scaling degraded faster for the 24-puzzle. 

Much of the previous literature on parallel search has focused on
maximizing the usage of all available CPU cores.
Our results indicate that  this approach can be suboptimal.
In fact, the best performance can be
obtained by keeping many of the available processors idle.  This is a
result of the prevalent architecture in current clusters, which are
composed of commodity, multicore shared-memory processors connected
via a high-speed network.  
There are two distinct factors which can lead to suboptimal performance when CPU core usage is maximized.
First, as we showed in Section \ref{sec:scale-nodes},
contention for local memory on each machine becomes a
bottleneck, so using all of the available cores on a machine results
in performance degradation.  
Automatically determining  the optimal number of cores to use per processing node in order to obtain the best tradeoff between computation and memory bandwidth is an area for future work.
Second, in parallel best-first search,
there is a tradeoff between the number of cores used per machine, and
the fragmentation of the local open list. Fragmentation of the open
list results in search overhead,
because the local open lists are not
aware of the current global best $f$-values.
Developing mechanisms which seek to reduce this search overhead is an area for future work.

Our comparison of HDA* with TDS showed that when HDA* had sufficient
memory to solve a problem, it significantly outperformed TDS on our
benchmark planning instances.  We showed that this performance gap was
due to the reexpansion of states by TDS.  
We observed that on problem instances where HDA* fails due to memory exhaustion, TDS either fails due to exceeding the time limit, or TDS eventually solves the instance but requires a long time.
Thus, we investigated a simple,
hybrid strategy which first executes HDA* until either the problem is
solved or until memory is exhausted. If HDA* fails, execute TDS
starting with a more informed initial threshold, provided by the failed HDA* run.
We showed that this hybrid strategy effectively combines the
advantages of both HDA* and TDS.

One particularly attractive feature of HDA* is its simplicity. Work
distribution and duplicate detection is done by a simple hash function, and there is no
complex load balancing mechanism. 
All communications are asynchronous. %
As far as we know, HDA* is the simplest parallelization for A* 
which achieves both load balancing and duplicate detection.
Simplicity is very important for parallel algorithms,
particularly for an algorithm that runs on multiple machines, as debugging a multi-machine, multicore algorithm is extremely challenging.
In some preliminary efforts to  implement a
distributed memory, work-stealing algorithm, we have found that it is
significantly more difficult to implement it correctly and efficiently
compared to HDA*.
Furthermore, using the standard MPI message passing library, the same
HDA* code can be recompiled and executed on a wide range of shared
memory and distributed clusters, making it simple to port HDA*. This
is in contrast to more complex algorithms which depend on specific
architectural features such as shared memory.

While our investigation of HDA* scaling reveals that there is clearly
room for improvement when the amount of parallel resources used
greatly exceeds the minimal required resources (i.e., $p > 8 \times
p_{min}$), we have shown that HDA* scales reasonably well across a
wide range of parallel platforms, including a single multicore machine, 
a commodity cluster, and two high-performance computing
clusters. Combined with its simplicity and portability, this suggests that 
HDA* should be considered a default, baseline algorithm for parallel
best-first search.

This paper focused on the search phase of problem
solving. Parallelization of the heuristic construction phase (e.g.,
computation of the abstraction heuristic table \cite{Helmert:07} and
pattern database generation \cite{KorfF02}) is another area for future
work.
A related avenue for future work is a more effective use of memory for heuristic tables.
Our current implementation of HDA* uses a single process per core. 
When executed 
on one or more (shared memory) multicore machines,
our implementation of HDA* executes as a
set of independent processes without sharing any memory resources
among cores that are on the same machine. This means that the memory
used for an abstraction heuristic in planning or for a pattern database
in 24-puzzle is
unnecessarily replicated $n$ times on an $n$-core machine.
We are currently
investigating a hybrid, distributed/shared memory implementation of
HDA* which eliminates this inefficiency.  One possible approach 
is to distribute work among machines
using hash-based distribution, but within a single machine incorporate
techniques such as speculative expansion that have been shown to scale well on a
shared memory environment~\cite{Burns:09}.

Finally, although efficiency
(with respect to runtime) is an important characteristic for parallel
search, the ability to solve difficult problems that cannot be solved
using fewer resources (because A* exhausts RAM, and other methods such as
IDA* on a single machine take too long) may be the most compelling reason to consider large scale parallel search on distributed memory clusters. 
Using up to 2400 cores and 10.5
terabytes of aggregate RAM, HDA* was able to compute optimal solutions
to larger planning benchmark problems than was previously possible on
a single machine.  
The increasing
pervasiveness of massive ``utility computing'' resources (such as
cloud computing services) means that further work is needed to develop and evaluate algorithms that can scale even further, to tens of thousands of cores and petabytes of RAM. Furthermore, a focus on solving problems using utility computing services which incur monetary usage costs requires the development of cost-effective utilization of resources. We have recently analyzed an iterative resource allocation policy to address this \cite{FKB12}.

\section*{Acknowledgments}

This research is supported by the JSPS Compview GCOE, 
the JST PRESTO program, and JSPS grants-in-aid for research.
Thanks to Malte Helmert for providing the Fast Downward code, and to Rich
Korf for providing his IDA* 24-puzzle solver, pattern database code, and puzzle
instances.
We thank the anonymous reviewers for their feedback.

\bibliography{ref}
\bibliographystyle{elsarticle-num}

\end{document}